\documentclass[11pt]{article}
\usepackage[margin=1in]{geometry}
\usepackage[backend=biber,style=numeric-comp,sorting=none]{biblatex}
\usepackage{amsmath,amssymb,amsfonts}
\usepackage{graphicx}
\graphicspath{{images/}{images/results/}}
\usepackage{subfig}
\usepackage{booktabs}
\usepackage{tabularx}
\usepackage{textcomp,nicefrac}
\usepackage{algorithm}
\usepackage{algpseudocode}

\usepackage{tikz}
\usepackage[section]{placeins}
\usetikzlibrary{arrows.meta,positioning,fit,calc,decorations.pathreplacing}
\usetikzlibrary{shapes.geometric}
\usepackage[hidelinks]{hyperref}

\newcommand{\keywords}[1]{\par\noindent\textbf{Keywords---} #1}

\begin{document}
\title{Graph Neural ODE Digital Twins for Control-Oriented Reactor Thermal-Hydraulic Forecasting Under Partial Observability}
\author{Akzhol Almukhametov$^{a}$, Doyeong Lim$^{a}$, Rui Hu$^{b}$, and Yang Liu$^{a}$\thanks{Corresponding author. E-mail: y-liu@tamu.edu}\\[0.4em]
\small $^{a}$Department of Nuclear Engineering, Texas A\&M University, College Station, TX 77843, USA\\
\small $^{b}$Argonne National Laboratory, Nuclear Science and Engineering Division, USA}

\date{}
\maketitle

\begin{abstract}
Real-time supervisory control of advanced reactors requires accurate forecasting of plant-wide thermal-hydraulic states, including locations where physical sensors are unavailable. Meeting this need calls for surrogate models that combine predictive fidelity, millisecond-scale inference, and robustness to partial observability. In this work, we present a physics-informed message-passing Graph Neural Network coupled with a Neural Ordinary Differential Equation (GNN-ODE) to addresses all three requirements simultaneously.

We represent the whole system as a directed sensor graph whose edges encode hydraulic connectivity through flow/heat transfer-aware message passing, and we advance the latent dynamics in continuous time via a controlled Neural ODE. A topology-guided missing-node initializer reconstructs uninstrumented states at rollout start; prediction then proceeds fully autoregressively. 
The GNN-ODE surrogate achieves satisfactory results for the system dynamics prediction. On held-out simulation transients, the surrogate achieves an average MAE of 0.91 K at 60 s and 2.18 K at 300 s for uninstrumented nodes, with $R^2$ up to 0.995 for missing-node state reconstruction. Inference runs at approximately 105 times faster than simulated time on a single GPU, enabling 64-member ensemble rollouts for uncertainty quantification.
To assess sim-to-real transfer, we adapt the pretrained surrogate to experimental facility data using layerwise discriminative fine-tuning with only 30 training sequences. The learned flow-dependent heat-transfer scaling recovers a Reynolds-number exponent consistent with established correlations, indicating constitutive learning beyond trajectory fitting. The model tracks a steep power change transient and produces accurate trajectories at uninstrumented locations.
\end{abstract}


\section{Introduction}
\label{sec:introduction}
The integration of variable renewable energy sources into the grid has elevated Small Modular Reactors (SMRs) and microreactors as critical assets for firm, flexible generation \cite{iea2022world, mignacca2020smr}. However, the lower power outputs of these designs impose tight economic constraints, driving a shift toward autonomous operation to reduce staffing costs and enable remote deployment \cite{decandido2024assessment, shirvan2023uo2}. To manage this autonomy safely, the nuclear community is adopting multi-layered supervisory architectures that decouple safety-critical protection from economic optimization, using constraint-enforcement algorithms such as Command Governors to keep the reactor within a defined ``Normal Operation Region'' \cite{ponciroli2024design, dave2023design, sun2024safe}. Recent studies have reinforced this trajectory through surrogate-assisted model predictive control \cite{na2006model} and deep reinforcement learning for reactor regulation \cite{chen2022deep}, underscoring the broader move toward optimization- and learning-based supervisory control. A common prerequisite across all such frameworks is access to fast and reliable dynamic models capable of forecasting plant behavior within the time budgets imposed by real-time control loops.

Effective supervisory control relies on Control-oriented Digital Twins (DTs) to forecast system behavior in real time. Recent frameworks have demonstrated DTs for validation and state inference \cite{stewart2025agn, liu2025development, mengyan2024current, ndum2024digital}, yet high-fidelity physics simulations remain computationally prohibitive for millisecond-level decision-making. Machine Learning (ML) surrogates have emerged to bridge this speed-accuracy gap: Physics-Informed Neural Networks, deep-learning-based model reduction, and related approaches have achieved orders-of-magnitude speedups for thermal-hydraulic transient modeling \cite{he2022deep, antonello2023physics, prantikos2023pinn}, enabling targeted demonstrations of autonomous reactor control through deep reinforcement learning and hybrid PID-neural network architectures \cite{tunkle2025nuclear, rigby2025reinforcement, lin2021development}.


Neural Ordinary Differential Equations (NODEs) offer a theoretically grounded route to close the gap between high-fidelity simulation and real-time decision making. Unlike discrete-time sequence models (e.g., RNN/GRU variants) that learn dynamics at a fixed sampling interval, NODEs parameterize the state derivative with a neural network and integrate it in continuous time \cite{chen2018neural, Neuromancer2023}. This makes the learned dynamics naturally invariant to the measurement rate and well suited to supervisory settings where sampling can be irregular or heterogeneous across sensors. In addition, the continuous-time formulation connects directly to Model Predictive Control (MPC) through adjoint-based sensitivities, enabling efficient gradient-based optimization over multi-step rollouts \cite{zhao2024safe, shen2025robust}.

Motivated by this perspective, our prior work \cite{lim2026autonomous} developed NODE surrogates for the same thermal-hydraulic testbed considered here and demonstrated accurate forecasting on held-out load-following and SCRAM transients with inference times compatible with real-time supervisory control. Nevertheless, the temporal advantages of NODEs were realized through a multilayer perceptron (MLP) parameterization that treated the sensor network as an unstructured vector, thereby neglecting the facility's known spatial connectivity.

In this work, we couple NODE-based temporal modeling with Graph Neural Networks (GNNs) \cite{scarselli2009gnn, kipf2016semi} to construct a GNN-ODE architecture that embeds hydraulic topology (pipes, valves, and heat exchangers) through physics-informed message passing. This hybrid design targets a central requirement for thermal-hydraulic autonomy: accurate multi-step forecasting that preserves physical connectivity, remains robust under variable-rate sensing, and stays differentiable for downstream trajectory optimization.

Graph-structured models have already shown value in nuclear applications, including whole-system digital twins \cite{liu2025development}, sensor fault detection and anomaly diagnosis \cite{liu2024graph, zhang2024spatial, chae2022graph, wang2025interpretability}, and in-core power distribution forecasting \cite{Lin2026STHDNCorePower}. However, these approaches uniformly rely on discrete-time temporal modules (CNN, RNN, or transformer blocks), which tie the learned dynamics to a fixed sampling interval and complicate integration with optimization-based control. We address this gap by pairing graph-based spatial propagation with a Neural ODE temporal backbone that learns continuous-time dynamics, naturally accommodates variable-rate and asynchronous sampling, and produces end-to-end differentiable rollouts compatible with gradient-based trajectory optimization and differentiable predictive control \cite{drgona2024}.

In our formulation, the reactor system is represented as a sensor graph whose nodes correspond to instrumented components and whose edges encode hydraulic connectivity. At each observation time, the model receives a partially observed state vector in which unavailable measurements (e.g., due to sensor outages, communication dropouts, or asynchronous sampling) are treated as masked inputs. Conditioned on the available sensors, the GNN-ODE propagates information over the graph and integrates the learned dynamics to produce multi-step forecasts and reconstructions of missing signals, enabling state prediction under partial observability. This reconstruction capability is also relevant to data-integrity settings in which a subset of measurements may be unreliable, and complements recent work on off-situ signal reconstruction under degraded telemetry \cite{Cao_off_situ_2021, Cao_off_situ_2026}.

This work contributes to the reactor instrumentation and controls domain by providing a surrogate architecture that functions as a virtual sensor array for inaccessible thermal-hydraulic states while delivering the inference speed required for closed-loop supervisory control. The main contributions are: (1) a physics-informed GNN-ODE surrogate that couples graph-based spatial propagation with continuous-time latent dynamics, embedding hydraulic topology through flow-aware message passing and learning interpretable constitutive relations directly from data; (2) a topology-guided missing-node initialization and autoregressive correction mechanism that reconstructs temperatures at uninstrumented plant locations from sparse sensor coverage; (3) experimental validation of sim-to-real transfer, in which the SAM-pretrained surrogate is adapted to physical facility data using discriminative fine-tuning and evaluated on a steep multi-step power transient with physically plausible reconstruction of uninstrumented nodes; and (4) efficient single-GPU inference enabling parallel ensemble rollouts for uncertainty quantification within real-time control horizons. To the best of our knowledge, this is the first demonstration of a graph-structured Neural ODE surrogate validated on experimental reactor-relevant facility data under partial observability.

The remainder of this manuscript is organized as follows. Section \ref{sec:methodology} presents the mathematical formulation of the proposed GNN-ODE architecture. Section \ref{sec:case_study} describes the thermal-hydraulic system representation and its high-fidelity SAM-based digital twin used that generated synthetic data used for the model's training. Section \ref{sec:results} reports performance metrics and evaluates predictive accuracy across a range of transient scenarios. Section \ref{sec:conclusions} concludes by summarizing the main findings and outlining directions for future work toward autonomous control implementation.

\section{Methodology}
\label{sec:methodology}

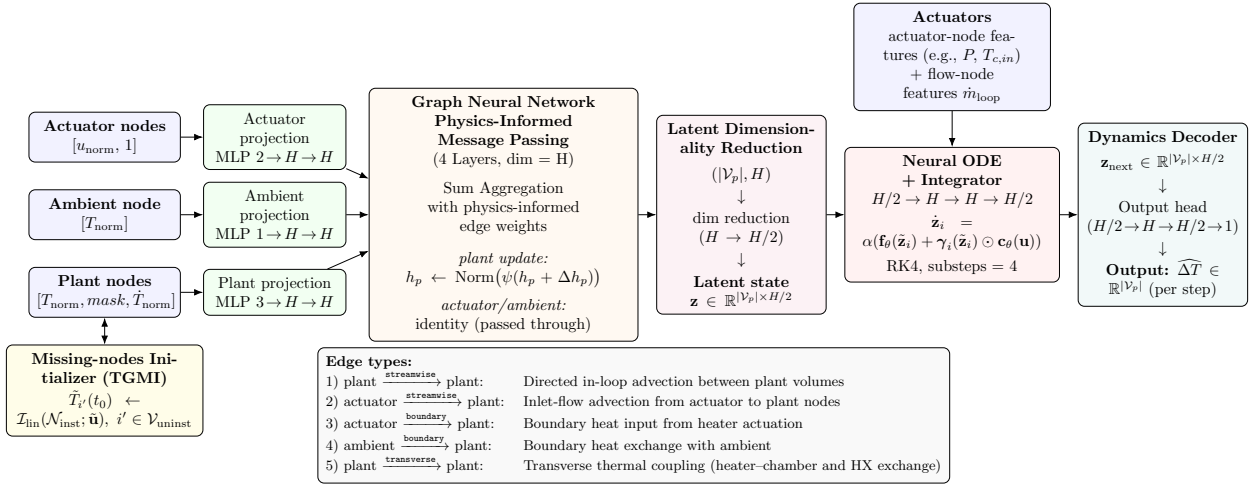
\begin{figure}[t]
\centering
\resizebox{\textwidth}{!}{%
\begin{tikzpicture}[
  font=\small,
  >=Latex,
  box/.style={draw, rounded corners, align=center, inner sep=5pt},
  io/.style={box, fill=blue!5, text width=3.0cm},
  proj/.style={box, fill=green!5, text width=2.8cm},
  gnn/.style={box, fill=orange!5, text width=5.6cm, minimum height=4.5cm},
  latent/.style={box, fill=purple!5, text width=3.4cm},
  odeblock/.style={box, fill=red!5, text width=4.4 cm},
  outbox/.style={box, fill=teal!5, text width=3.4cm},
  legend/.style={draw, rounded corners, inner sep=5pt, font=\footnotesize, align=left, fill=gray!5}
]

\node[io] (actin) {\textbf{Actuator nodes}\\$[u_{\mathrm{norm}},\, 1]$};
\node[io, below=0.6cm of actin] (ambin) {\textbf{Ambient node}\\$[T_{\mathrm{norm}}]$};
\node[io, below=0.6cm of ambin] (plantin) {\textbf{Plant nodes}\\$[T_\mathrm{norm}, mask, \dot T_{\mathrm{norm}}]$};

\node[box, fill=yellow!10, text width= 4.0 cm, below=0.6cm of plantin] (TGI) {\textbf{Missing-nodes Initializer (TGMI)}\\[2pt] $\tilde{T}_{i'}(t_0) \leftarrow \mathcal{I}_{\mathrm{lin}}(\mathcal{N}_{\mathrm{inst}}; \tilde{\mathbf{u}}),\ i'\in\mathcal{V}_{\mathrm{uninst}}$};
\draw[<->, line width=0.6pt] (TGI) -- (plantin);

\node[proj, right=0.5cm of actin] (actproj) {Actuator projection\\MLP $2\!\rightarrow\!H\!\rightarrow\!H$};
\node[proj, right=0.5cm of ambin] (ambproj) {Ambient projection\\MLP $1\!\rightarrow\!H\!\rightarrow\!H$};
\node[proj, right=0.5cm of plantin] (plantproj) {Plant projection\\MLP $3\!\rightarrow\!H\!\rightarrow\!H$};

\draw[->, line width=0.6pt] (actin) -- (actproj);
\draw[->, line width=0.6pt] (ambin) -- (ambproj);
\draw[->, line width=0.6pt] (plantin) -- (plantproj);

\node[gnn, right=0.5cm of ambproj] (gnnlayer) {\textbf{Graph Neural Network\\ Physics-Informed Message Passing}\\(4 Layers, dim = H)\\\vspace{6pt}
Sum Aggregation with physics-informed\\edge weights\\\vspace{8pt}
\textit{plant update:}
\\$h_p \leftarrow \mathrm{Norm}\big(\psi(h_p + \Delta h_p)\big)$\\\vspace{6pt}
\textit{actuator/ambient:}\\identity (passed through)};

\draw[->, line width=0.6pt] (actproj) -- (gnnlayer.165);
\draw[->, line width=0.6pt] (ambproj) -- (gnnlayer.180);
\draw[->, line width=0.6pt] (plantproj) -- (gnnlayer.195);

\node[latent, right=0.4cm of gnnlayer] (latent) {\textbf{Latent Dimensionality Reduction}\\\vspace{4pt}
$(|\mathcal{V}_p|,H)$\\\vspace{2pt}$\downarrow$\\\vspace{2pt}
dim reduction ($H\!\rightarrow\!H/2$)\\\vspace{2pt}$\downarrow$\\\vspace{2pt}
\textbf{Latent state} $ \textbf{z} \in\mathbb{R}^{|\mathcal{V}_p| \times H/2}$};

\draw[->, line width=0.6pt] (gnnlayer) -- (latent);

\node[odeblock, right=0.4cm of latent] (ode) {\textbf{Neural ODE + Integrator}\\ $H/2 \!\rightarrow\! H\!\rightarrow\!H\!\rightarrow\!H/2$ \\\vspace{2pt}
$\dot{\mathbf{z}}_i = \alpha\!\left(\mathbf{f}_\theta(\tilde{\mathbf{z}}_i) + \boldsymbol{\gamma}_i(\tilde{\mathbf{z}}_i)\odot \mathbf{c}_\theta(\mathbf{u})\right)$ \\ \vspace{4pt}
RK4, substeps = 4};

\node[io, text width=4.0cm, above=0.8cm of ode] (u) {\textbf{Actuators}\\actuator-node features (e.g., $P,\,T_{c,in}$)\\+ flow-node features $\dot{m}_{\mathrm{loop}}$};
\draw[->, line width=0.6pt] (u) -- (ode);
\draw[->, line width=0.6pt] (latent) -- (ode);

\node[outbox, right=0.4cm of ode] (outhead) {\textbf{Dynamics Decoder}\\\vspace{4pt}
$\mathbf{z}_{\mathrm{next}}\in\mathbb{R}^{|\mathcal{V}_p| \times H/2}$\\\vspace{2pt}$\downarrow$\\\vspace{2pt}
Output head\\($H/2\!\rightarrow\!H\!\rightarrow\!H/2\!\rightarrow\!1$)\\\vspace{2pt}$\downarrow$\\\vspace{2pt}
\textbf{Output:} $\widehat{\Delta T}\in\mathbb{R}^{|\mathcal{V}_p|}$ (per step)};

\draw[->, line width=0.6pt] (ode) -- (outhead);

\node[legend, below= 3.8cm of $(actin)!0.5!(outhead)$] (edges) {
  \textbf{Edge types:}\\
  \begin{tabular}{@{}ll@{}}
  1) plant $\xrightarrow{\texttt{streamwise}}$ plant: & Directed in-loop advection between plant volumes\\
  2) actuator $\xrightarrow{\texttt{streamwise}}$ plant: & Inlet-flow advection from actuator to plant nodes \\
  3) actuator $\xrightarrow{\texttt{boundary}}$ plant: & Boundary heat input from heater actuation\\
  4) ambient $\xrightarrow{\texttt{boundary}}$ plant: & Boundary heat exchange with ambient\\
  5) plant $\xrightarrow{\texttt{transverse}}$ plant: & Transverse thermal coupling (heater--chamber and HX exchange)
  \end{tabular}
};

\end{tikzpicture}%
}
\caption{Architecture of the proposed physics-informed GNN-ODE surrogate for thermal-hydraulic forecasting under partial observability. Plant-node features are represented as $[T_\mathrm{norm}, mask, \dot T_{\mathrm{norm}}]$, and a topology-guided missing-node initializer (TGMI) provides initial estimates for uninstrumented nodes. Type-specific MLP encoders map plant, actuator, and ambient inputs to a shared hidden space $H$, followed by a three-layer GNN encoder in which plant states are updated via physics-informed message passing while actuator and ambient states are propagated unchanged. A controlled latent Neural ODE, integrated with Runge Kutta 4-th order (RK4), advances the latent state, and the decoder predicts node-wise temperature increments $\widehat{\Delta T}\in\mathbb{R}^{|\mathcal{V}_p|}$, where $|\mathcal{V}_p|$ is a number of plant nodes. The control input $u$ is assembled from actuator-node and flow-node features.}
\label{fig:architecture}
\end{figure}

We propose a hybrid surrogate modeling architecture that integrates a Graph Neural Network (GNN) for spatial topology with a Neural Ordinary Differential Equation (NODE) for continuous-time dynamics. This design extends the MLP-based NODE surrogate developed in our prior work \cite{lim2026autonomous} for the same testbed, which parameterized the state derivative as a function of a flat sensor vector and demonstrated accurate single-step forecasting but did not encode the facility's hydraulic connectivity. The schematic of the resulting GNN-ODE model is shown in Figure~\ref{fig:architecture}. This section details the graph construction, the physics-informed message passing mechanisms, and the temporal integration scheme.

\subsection{Spatial Modeling via Physics-Informed Graph Neural Networks}
\label{sec:spatial_modeling}

To capture the non-Euclidean topology of the thermal-hydraulic facility, we represent the system as a directed heterogeneous graph $\mathcal{G}=(\mathcal{V},\mathcal{E})$. The node set $\mathcal{V}$ is partitioned into three types: \textit{Plant Nodes} ($\mathcal{V}_p$), representing physical control volumes; \textit{Actuator Nodes} ($\mathcal{V}_a$), representing controllable boundary conditions; and \textit{Ambient Nodes} ($\mathcal{V}_{amb}$). We further partition plant nodes according to sensor availability into instrumented nodes $\mathcal{V}_{\mathrm{inst}}\subseteq\mathcal{V}_p$ and uninstrumented (missing) nodes $\mathcal{V}_{\mathrm{uninst}}\subseteq\mathcal{V}_p$.
The edge set $\mathcal{E}$ comprises three relation types that mimic the dominant transport mechanisms in the studied facility: advection-type edges, convection--conduction-type edges, and actuator/boundary-condition edges.

\subsubsection{Feature Normalization and Typed Projections}
\label{subsec:Norm&Encoding}

Neural networks train more reliably when inputs have comparable scales. In our case, the raw signals vary widely in magnitude, with temperatures around $300$--$400$~K, heater power in the range $0$--$15 000$~W, and mass flow rates around $0.01$--$0.15$~kg/s.

For plant nodes, the input channel is explicitly defined as
\begin{equation}
\mathbf{x}^{\mathrm{plant}}_i = [T_{i\, \mathrm{norm}},\, m_i,\, \dot{T}_{i\,\mathrm{norm}}],
\end{equation}
where $T_{i\,\mathrm{norm}}$ is standardized using node-specific statistics,
$T_{i\,\mathrm{norm}}=(T_i-\mu_{T_i})/\sigma_{T_i}$; $m_i\in\{0,1\}$ denotes the observability mask (see Subsection~\ref{subsec:partial_obs}).
$\dot{T}_{i\,\mathrm{norm}}=(dT_i/dt)/s_i$ is a temporal-gradient feature that acts as a compact ``thermal-momentum'' embedding, where $s_i$ denotes the node-specific standard deviation computed once from the full available dataset. This feature resolves a practical ``cold-start'' ambiguity: at nodes far from the heat source, transport-induced delays can make a single temperature snapshot insufficient to determine whether a node is heating or cooling. Estimating $\dot{T}_{i\,\mathrm{norm}}$ from a short measurement history provides this directional information without introducing recurrent memory, preserving the Markovian structure of the dynamics model.

Actuator nodes corresponding to reactor power ($P$) and coolant inlet temperature ($T_{c,in}$) (i.e., the tertiary-loop inlet) are encoded as two-dimensional vectors \([u_{\mathrm{norm}},\,1]\), where \(u_{\mathrm{norm}}\) is the normalized control input and the second entry is a constant bias channel. The bias channel provides an always-active offset that enables the actuator projection layers to capture control-independent baseline effects, yielding a numerically stable encoding of affine actuator effects in the shared latent space.

For the ambient node, we use the normalized scalar temperature input $\mathbf{x}^{\mathrm{amb}}=[T_{\mathrm{amb\,norm}}]$.

In addition to node-wise features, we include a global operating-condition vector, \(\dot{\mathbf m}_{\mathrm{loop}}\in\mathbb{R}^{n_{\mathrm{loop}}}\), containing loop-level mass-flow rates. This global input conditions both the latent dynamics and flow-dependent interaction strengths across the graph; \(n_{\mathrm{loop}}\) is determined by the facility topology.

Power actuator input and flow rates are linearly mapped to $[0,1]$ using the physical operating bounds, while coolant inlet node and ambient node are normalized as plant node temperatures. This preconditioning prevents the encoder MLPs from being dominated by large-magnitude channels (especially power) and improves sensitivity to smaller thermal variations. 

To align heterogeneous node features in a common latent space, the typed inputs defined above are projected into a shared hidden dimension $H$ using dedicated, type-specific MLPs (Fig.~\ref{fig:architecture}). For each node type $r$, the input $x_i^{(r)}$ is mapped by a two-layer perceptron:
\begin{equation}
    h_i^{(r)} = W_2^{(r)}\,\phi\!\left(W_1^{(r)}x_i^{(r)} + b_1^{(r)}\right) + b_2^{(r)},
\end{equation}
where $\phi(\cdot)$ is Gaussian Error Linear Unit (GELU) \cite{GELU}. This formulation embeds all node types in the same hidden space $H$ and provides smooth nonlinear gating with stable optimization in continuous-valued thermal regimes.

\subsubsection{Node Representation and Partial Observability Accommodation}
\label{subsec:partial_obs}

Under partial observability, the continuous-time ODE solver still requires a fully specified initial state $\mathbf{x}(t_0)$, even when some plant locations are uninstrumented. We therefore combine an observability-aware input representation with a two-stage initialization-and-correction procedure.

Concretely, each node feature vector $\mathbf{x}_i$ is augmented with a binary \textit{observability mask} $m_i \in \{0,1\}$ indicating whether a measurement is available. For instrumented nodes, $m_i=1$ and the measured temperature is supplied directly. For uninstrumented nodes, we set $m_i=0$ and initialize the missing temperature via a lightweight \textit{Topology-Guided Missing-nodes Initializer} (TGMI) fit offline on the full set of high-fidelity simulation trajectories used in this work. For each $i'\in\mathcal{V}_{\mathrm{uninst}}$, we define the regressor set $\mathcal{N}_{\mathrm{inst}}(i')$ strictly as its 1-hop instrumented neighbors in $\mathcal{G}$ (i.e., nodes directly connected to $i'$ via streamwise transport or transverse exchange edges, including measured boundary/actuator temperatures when applicable). We augment these neighbor features with an expanded global context vector $\tilde{\mathbf{u}}$ (containing heater power, loop flow rates and their quadratic combinations), normalize all regressors, and fit an ordinary least-squares model. The resulting initialization is summarized in the linear form below, providing a numerically stable seed for $\mathbf{x}(t_0)$ that is refined in the second stage:

\begin{equation}
\label{eq:tgmi}
    \hat{T}_{i'} = \sum_{k \in \mathcal{N}_{\mathrm{inst}}(i')} w_{i'k} \, T_k + \mathbf{v}_{i'}^{\top} \tilde{\mathbf{u}} + b_{i'},\quad i'\in\mathcal{V}_{\mathrm{uninst}}
\end{equation}
where the coefficients \(\{w_{i'k}\}, \mathbf{v}_{i'}\), and \(b_{i'}\) are estimated by solving an ordinary least-squares (OLS) problem for each \(i' \in \mathcal{V}_{\mathrm{uninst}}\), i.e., minimizing the sum of squared residuals between measured and predicted temperatures over the training data.

By facility design, each uninstrumented internal volume has at least one 1-hop instrumented neighbor, ensuring the initializer is always well-defined. We select ordinary least-squares regression for its negligible inference cost and sufficient precision for seeding the rollout. In implementation, the TGMI operates as a preprocessing step before rollout rather than as an internal GNN module, although it shares the graph topology to identify neighboring nodes.

The second stage occurs during the GNN-ODE rollout and constitutes the surrogate's physically meaningful recovery mechanism. Once the initial estimates seed the state, graph message passing propagates thermal information through the known hydraulic connectivity along the predefined streamwise transport and transverse exchange edges. Continuous-time integration then progressively refines the imputed values as the trajectory evolves. In this way, the initializer serves merely as a mathematical starting point, while the learned dynamics correct the coarse estimates by enforcing the energy-balance structure encoded in the graph. This mechanism supports reliable inference at sparsely instrumented locations and can be extended to maintain consistent state estimates when measurements are missing, delayed, or corrupted.

\subsubsection{Physics-Informed Message Passing as Energy Balance}
\label{subsec:PI Message Passing}

Spatial interactions are modeled with a heterogeneous message-passing scheme that mirrors a discretized energy-balance formulation: each node represents a lumped control volume, and directed edges represent energy transport pathways (advection along the hydraulic direction and transverse exchange across a thermal interface). To keep the surrogate low order, we do not explicitly model momentum conservation. Instead, each loop mass flow rate is treated as a measured global variable, collected in $\mathbf{\Phi}$, and assumed spatially uniform within the loop during a timestep. These flow variables play a \textit{triple role} in the architecture: they (i) scale advective enthalpy transport, (ii) modulate effective heat-transfer coefficients, and (iii) enter the temporal model as exogenous inputs to the latent ODE. 

At layer $l$, each edge $j\rightarrow i$ produces a relation-specific message $\mathcal{M}_r(\cdot)$ that can be interpreted as a learned approximation of an energy flux contribution. Messages are then aggregated by summation to form a net ``incoming energy'' signal at node $i$, and combined with the prior node embedding through a residual update:

\begin{equation}
\label{eq: Message Aggregation}
    \mathbf{h}_i^{(l+1)} = \sigma \left( \mathbf{h}_i^{(l)} + \sum_{r \in \mathcal{R}} \sum_{j \in \mathcal{N}_r(i)} \mathcal{M}_r(\mathbf{h}_i^{(l)}, \mathbf{h}_j^{(l)}, \mathbf{e}_{ji}) \right)
\end{equation}

In the implemented architecture, this residual message-passing update is applied to plant nodes, while actuator and ambient node embeddings are propagated unchanged across the layer. Here, \(\mathbf{e}_{ji}\) denotes the edge-attribute vector for edge \(j\!\to\! i\), containing relation-dependent physical metadata that modulate message strength. In practice, the post-aggregation operator \(\sigma(\cdot)\) corresponds to a normalization-activation-regularization block (Layer Normalization, GELU, and dropout), rather than a single pointwise nonlinearity.

Edge flux terms are scaled using destination-node volume normalization, yielding consistent edge-rate magnitudes under the assumption of approximately constant fluid properties (density and heat capacity). In this formulation, each node volume is approximated from the fluid volume of the corresponding piping segment; edge attributes \(\mathbf{e}_{ji}\) include these normalized rate terms together with relation-specific indicators (e.g., boundary/conduction flags and flow-dependent modifiers).  

While this representation is primarily fluid-volume based, metal thermal inertia is not entirely ignored: for selected components (notably the heater-associated node), an effective volume correction is applied using a steel-to-water volumetric heat-capacity ratio. More generally, the same normalization framework can be extended to explicit node-wise thermal-mass definitions when higher-fidelity solid capacitance modeling is required.

The aggregated message in \eqref{eq: Message Aggregation} comprises physically motivated components that correspond to the dominant energy transport mechanisms in the facility:

\textit{(i) Streamwise transport / advection edges.} Messages on \textit{streamwise-transport} edges represent flow-driven enthalpy transport along the piping network. We explicitly provide the loop mass flow rate $\dot{m}_{\mathrm{loop}}$ (or its normalized counterpart $\phi$) as a measured input that scales the message magnitude. The remaining dependence on local state is learned, so that the message can represent both linear advection and unmodeled mixing/dispersion effects:
\begin{equation}
    \mathcal{M}_{\mathrm{stream}}(j\rightarrow i) = \dot{m}_{\mathrm{loop}} \, f_{\mathrm{stream}}\!\left(\mathbf{h}_j^{(l)},\mathbf{h}_i^{(l)}\right),
\end{equation}
where $f_{\mathrm{stream}}(\cdot)$ is a neural function shared across all advection edges of the same type. This structure preserves the physical scaling with flow while allowing the model to learn the effective relationship between upstream/downstream embeddings and advective heat transport.

\textit{(ii) Transverse exchange / convection + conduction-like transfer edges.} Messages on \textit{transverse exchange} edges represent lumped heat transfer between distinct thermal domains, including (a) coupling across heat exchanger walls (primary--secondary exchange) and (b) exchange between solids and fluids (e.g., heat exchanger walls and heater rods to coolant). Each edge is interpreted as an effective series thermal resistance, summarized by a conductance $\mathrm{UA}_{\mathrm{eff}}$. Rather than hard-coding correlations, we condition a learned conductance model on physical context features (component-type flags and local flows):
\begin{equation}
    \mathcal{M}_{\mathrm{trans}}(j\rightarrow i) = \mathrm{UA}_{\mathrm{eff}} (\mathbf{h}_j^{(l)},\mathbf{h}_i^{(l)},\mathbf{c}_{ji}) \, \Delta T_{ji},
\end{equation}
where $\Delta T_{ji}$ is represented implicitly through the embeddings and $\mathbf{c}_{ji}$ collects defined context (e.g., heat-exchanger/heater flags and source/destination flows).

Beyond low prediction error, the proposed physics-informed message passing learns physically consistent structure directly from data. Instead of hard-coding empirical heat-transfer correlations, we parameterize the effective conductance ($\mathrm{UA}_{\mathrm{eff}}$) with a differentiable scaling in which local heat-transfer coefficients are learned as functions of flow. Specifically, we assume $HTC \propto \dot{m}_{\mathrm{loop}}^{\alpha}$ and treat $\alpha$ as a trainable exponent. The resulting conductance is computed via a symmetric (harmonic-mean) coupling of source and destination flow scales:
\begin{equation}
\label{eq: HTC}
    S_{src} = (\phi_{src} + \epsilon)^{\alpha_r}, \quad S_{dst} = (\phi_{dst} + \epsilon)^{\alpha_r}
\end{equation}
\begin{equation}
    \mathrm{UA}_{\mathrm{eff}} \propto \frac{2 \cdot S_{src} \cdot S_{dst}}{S_{src} + S_{dst}}
\end{equation}

Crucially, the exponent $\alpha_r$ is learned independently for different component types, allowing the model to discover physically interpretable correlations.

\textit{(iii) Actuators and Boundary Conditions edges}. Finally, we introduce actuators and boundary conditions edges to represent energy source terms (e.g., Heater) and environmental heat losses. The Heater Power actuator node connects directly to the Heater Rod node. After normalization and encoding, this edge injects the heat generation term into the solid fuel approximation. 

To account for imperfect insulation, each plant node is connected to a global Ambient Node (fixed at $T_{amb} = 293.15$~K). The corresponding edge learns a component-specific conductance parameter $U_{loss}$, modeling parasitic heat rejection to the room as $\dot{Q}_{loss} = U_{loss}A\,(T_{node}-T_{amb})$.

In summary, all the messages are aggregated in Eq.~\ref{eq: Message Aggregation}, such that each plant-node embedding is updated through a residual transformation as:

\begin{equation}
\Delta\mathbf{h}^{(l)}_{p,i} = \sum_{r \in \mathcal{R}} \sum_{j \in \mathcal{N}_r(i)} \mathcal{M}_r(\mathbf{h}_i^{(l)}, \mathbf{h}_j^{(l)}, \mathbf{e}_{ji})
\end{equation}

\subsection{Temporal Dynamics via Neural Ordinary Differential Equations}
\label{sec:temporal_dynamics}

\subsubsection{Latent Continuous-Time Dynamics}

To model temporal evolution, we treat post-aggregation plant-node embeddings as inputs to a controlled continuous-time latent dynamical system. After the final GNN layer (\(L\)), which completes physics-informed graph propagation, each plant node is represented by \(\mathbf{h}^{(L)}_{p,i}\in\mathbb{R}^{H}\). This representation is then projected through a linear bottleneck to obtain a lower-dimensional latent state:
\begin{equation}
\begin{aligned}
\mathbf{z}_i(t) &= \mathbf{W}_{z}\mathbf{h}^{(L)}_{p,i}+\mathbf{b}_{z},\qquad \\
\mathbf{z}_i(t) &\in\mathbb{R}^{d_z},\quad
\mathbf{Z}(t)\in\mathbb{R}^{|\mathcal{V}_p|\times d_z}
\end{aligned}
\end{equation}
where \(d_z\) is the latent (reduced) dimension after bottleneck projection, which in our case equals to half of higher dimension ($d_z = H/2$). This defines an encoder--dynamics--readout pipeline that decouples spatial interaction modeling (graph propagation) from temporal evolution (ODE integration), which is beneficial under nonuniform sampling intervals.

We employ this compression--expansion (hourglass) latent bottleneck to enforce compact, dynamically relevant representations and improve generalization by reducing overparameterization; empirically, it outperformed architectures that maintain a uniformly high-dimensional state throughout.

The latent dynamics are parameterized as
\begin{equation}
\frac{d\mathbf{z}_i}{dt}
=
\alpha\!\left(
\mathbf{f}_\theta(\tilde{\mathbf{z}}_i)
+
\boldsymbol{\gamma}_i(\tilde{\mathbf{z}}_i)\odot \mathbf{c}_\theta(\mathbf{u})
\right),
\end{equation}
where \(\tilde{\mathbf{z}}_i\) is the layer-normalized latent state, \(\mathbf{f}_\theta\) is a two-hidden-layer nonlinear map (LayerNorm + SiLU \cite{silu} per hidden block), \(\mathbf{c}_\theta(\mathbf{u})\) is a learned projection of global controls \(\mathbf{u}\), \(\boldsymbol{\gamma}_i=\sigma(\mathbf{W}\tilde{\mathbf{z}}_i+\mathbf{b})\) is an elementwise gate, and \(\alpha\) is a learned scalar controlling derivative magnitude.

\subsubsection{Differentiable RK4 Integration and Training}

The initial-value problem is advanced with a differentiable fourth-order Runge--Kutta (RK4) scheme using optional internal substeps within each physical interval \(\Delta t\). With \(S\) RK4 substeps per interval, the internal step size is \(\delta=\Delta t/S\). Here, \(S\) is a tunable hyperparameter; in our case study, we use \(S=4\). For each substep,
\begin{equation}
\left\{
\begin{aligned}
\mathbf{k}_1 &= \mathbf{F}(\mathbf{z},\mathbf{u}), \\
\mathbf{k}_2 &= \mathbf{F}(\mathbf{z}+\tfrac{\delta}{2}\mathbf{k}_1,\mathbf{u}), \\
\mathbf{k}_3 &= \mathbf{F}(\mathbf{z}+\tfrac{\delta}{2}\mathbf{k}_2,\mathbf{u}), \\
\mathbf{k}_4 &= \mathbf{F}(\mathbf{z}+\delta\mathbf{k}_3,\mathbf{u}),
\end{aligned}
\right.
\end{equation}
\begin{equation}
\mathbf{z}\leftarrow
\mathbf{z}+\frac{\delta}{6}\left(\mathbf{k}_1+2\mathbf{k}_2+2\mathbf{k}_3+\mathbf{k}_4\right)
\end{equation}
Here, \(\mathbf{F}(\mathbf{z},\mathbf{u})\) denotes the learned latent vector field (i.e., the ODE right-hand side \(\mathrm{d}\mathbf{z}/\mathrm{d}t\)) evaluated from the current latent state and control inputs at each RK4 stage.

The integrated latent state \(\mathbf{Z}(t+\Delta t)\) is mapped to node-level outputs by a node-wise multilayer decoder (\(H/2 \!\to\! H \!\to\! H/2 \!\to\! 1\)) with GELU activations between linear layers. This expansion-contraction readout translates compact latent dynamics into physically meaningful temperature updates. The intermediate hidden expansion allows the decoder to model higher-order interactions not linearly recoverable from the bottleneck state. The resulting scalar output at each plant node is interpreted as the stepwise temperature increment \(\widehat{\Delta T}_i\) over \([t,t+\Delta t]\). Thus, the model does not directly output absolute temperature; absolute trajectories are recovered recursively via 
\begin{equation}
    \hat{T}_i(t+\Delta t)=\hat{T}_i(t)+\widehat{\Delta T}_i
\end{equation}


\begin{algorithm}
\small
\caption{Unified Training Procedure (Warm-Up $K{=}1$ $\rightarrow$ Curriculum BPTT $K{>}1$, scalar node state)}
\label{alg:unified_training_f1}
\begin{algorithmic}[1]
\Require \parbox[t]{0.78\linewidth}{\raggedright Trajectory window $\{T^{true}_{t-1:t+K_{\max}},\,u_{t:t+K_{\max}-1},\,\Delta t_{t:t+K_{\max}-1}\}$}
\Require \parbox[t]{0.78\linewidth}{\raggedright Graph $\mathcal{G}$; uninstrumented-node set $\mathcal{V}_{\mathrm{uninst}}$}
\Require \parbox[t]{0.78\linewidth}{\raggedright TGMI $\mathcal{I}_{\mathrm{lin}}(i,\cdot,\mathcal{G})$ for $i\in\mathcal{V}_{\mathrm{uninst}}$}
\Require Per-node rate scale $\sigma_r\in\mathbb{R}^{N}$
\Require \parbox[t]{0.78\linewidth}{\raggedright Horizon schedule $K(e)$; base TF schedule $p_{\mathrm{base}}(e)\downarrow 0$}
\Require RK internal substeps $S$ (used value: $S{=}4$)
\Ensure Updated parameters $\theta$

\For{epoch $e=1,\dots,E$}
    \State $K\gets K(e)$; if $e\le E_{\mathrm{warm}}$, set $K\gets 1$
    \State $p_{\mathrm{TF}}(e)\gets \mathrm{clip}\!\big(p_{\mathrm{base}}(e)+\Delta p_{\mathrm{bump}}(e),0,1\big)$

    \State $T^{roll}_0\gets T^{true}_t$, \quad $T^{roll}_{-1}\gets T^{true}_{t-1}$
    \For{$i\in\mathcal{V}_{\mathrm{uninst}}$}
        \State $T^{roll}_0[i]\gets \mathcal{I}_{\mathrm{lin}}(i,T^{roll}_0,\mathcal{G})$
        \State $T^{roll}_{-1}[i]\gets \mathcal{I}_{\mathrm{lin}}(i,T^{roll}_{-1},\mathcal{G})$
    \EndFor

    \State $\mathcal{L}\gets 0$
    \For{rollout step $\tau=0,\dots,K-1$}
        \State $\Delta t_\tau \gets \Delta t_{t+\tau}$
        \State Encode $(T^{roll}_{\tau},T^{roll}_{\tau-1},u_{t+\tau})$ to latent $z_{\tau}$
        \State $z_{\tau+1} \gets \mathrm{RK4}_S\!\big(f_\theta, z_\tau, u_{t+\tau}, \Delta t_\tau\big)$

        \State $\Delta \hat T_\tau \gets \mathrm{Decoder}_\theta(z_{\tau+1})$
        \State $\hat r_\tau \gets \Delta \hat T_\tau/\Delta t_\tau$
        \State $\Delta T_\tau \gets T^{true}_{t+\tau+1}-T^{true}_{t+\tau}$
        \State $r_\tau \gets \Delta T_\tau/\Delta t_\tau$
        \State $\mathcal{L}\gets \mathcal{L}+\frac{1}{K}\left\|(\hat r_\tau-r_\tau)\oslash \sigma_r\right\|_2^2$

        \State $\hat T_{\tau+1}\gets T^{roll}_\tau+\Delta \hat T_\tau$
        \If{$\tau<K-1$}
            \State Draw one Bernoulli at rollout step $\tau$: $b_\tau\sim \mathrm{Bernoulli}(p_{\mathrm{TF}}(e))$
            \State $T^{in}_{\tau+1}\gets b_\tau\,T^{true}_{t+\tau+1}+(1-b_\tau)\,\hat T_{\tau+1}$
            \For{$i\in\mathcal{V}_{\mathrm{uninst}}$}
                \State $T^{in}_{\tau+1}[i]\gets \mathcal{I}_{\mathrm{lin}}(i,T^{in}_{\tau+1},\mathcal{G})$
            \EndFor
            \State $T^{roll}_{\tau+1}\gets T^{in}_{\tau+1}$
        \Else
            \State $T^{roll}_{\tau+1}\gets \hat T_{\tau+1}$
        \EndIf
    \EndFor

    \State $\theta\gets \mathrm{Optimizer}\!\left(\theta,\nabla_\theta \mathcal{L}\right)$
\EndFor

\State \textbf{Validation/Testing:} set $p_{\mathrm{TF}}=0$ and run fully autoregressive rollout.
\end{algorithmic}
\end{algorithm}

\subsection{Training Strategy and Loss Formulation}
\label{sec:training_strategy}

For supervision in variable-step data, these increments are converted to rate predictions as \(\widehat{r}_i=\widehat{\Delta T}_i/\Delta t\) (and compared against \(r_i=\Delta T_i/\Delta t\), optionally after per-node scaling), which keeps the objective consistent across nonuniform sampling intervals.

All components in this pipeline are differentiable, including latent projection, graph-based spatial encoding, controlled vector-field evaluation, every RK4 stage/substep, and the final readout. Consequently, gradients are propagated end-to-end through the full unrolled temporal computation graph (backpropagation through time), so the loss at later prediction steps updates both the spatial message-passing parameters and the continuous-time latent dynamics jointly.

In this work, gradients are computed by direct automatic differentiation through the explicit fixed-step RK4 updates, rather than by an adjoint-sensitivity formulation. This choice preserves a fully transparent computational graph across all intermediate stages and substeps, which is advantageous for training stability and implementation simplicity in the present controlled-dynamics setting.


To ensure the surrogate model is robust to long-horizon rollouts while maintaining training stability, we employ a curriculum-based strategy utilizing Backpropagation Through Time (BPTT). This approach progressively transitions the model from learning instantaneous gradients to optimizing full trajectory dynamics. 

We adopt a scheduled-sampling \textit{Curriculum Learning with BPTT} \cite{bengio2015scheduled} to bridge single-step supervision and long-horizon rollout optimization. At each rollout step during training, the model receives either the ground-truth state or its own previous prediction, selected by per-step Bernoulli scheduled sampling. The teacher-forcing probability is initialized at 0.3 and decayed to 0 over epochs, progressively matching the fully autoregressive deployment regime while mitigating exposure bias; validation is conducted fully autoregressively (without teacher forcing). To further stabilize curriculum transitions, whenever the unroll horizon is increased, the teacher-forcing probability is temporarily increased by \(+0.15\) and then linearly decayed over the next 10 epochs. To balance stability and long-horizon accuracy, we use a dynamic unrolling schedule:

\begin{enumerate}
    \item \textbf{Phase 1 (Warm-up):} The model is trained with a prediction horizon of $k=1$. This mirrors "Teacher Forcing," where the model receives the ground truth at every step, allowing it to rapidly learn the basic topological heat transfer relations.
    \item \textbf{Phase 2 (Sequence Extension):} We progressively increase the rollout horizon to up to 64 steps. In this regime, the model's own predictions $\hat{\mathbf{x}}_{t+1}$ are fed back as inputs for subsequent steps. Gradients are propagated through the entire unrolled sequence (BPTT), directly penalizing the compounding errors that characterize drift in dynamical systems.
\end{enumerate}

The training objective is formulated as a normalized, mask-aware Mean-Squared Error (MSE) in rate space. For a trajectory of \(K\) steps, we first convert per-step temperature increments to rates, \(\hat r_{i,k}=\Delta \hat T_{i,k}/\Delta t_k\) and \(r_{i,k}=\Delta T_{i,k}^{true}/\Delta t_k\), and optimize the following loss function over the plant nodes $\mathcal{V}_p$:
\begin{equation}
    \mathcal{L} = \frac{1}{K \sum_{i\in\mathcal V_p} m_i} \sum_{k=1}^{K}\sum_{i\in\mathcal V_p} m_i \left(\frac{\hat r_{i,k}-r_{i,k}}{\sigma_i}\right)^2,  
\end{equation}
where $m_i \in \{0, 1\}$ represents the observability mask for node $i$, and \(\sigma_i\) is the standard deviation of the target temperature-change rate at node \(i\), computed from the training data as \(\sigma_i = \operatorname{Std}_{\text{train}}(\Delta T_i/\Delta t)\). This standardization ensures stable gradient contributions across nodes with vastly different thermal dynamics. 

This mask-aware formulation allows the loss to dynamically adapt to the two distinct training phases of the surrogate model.

\textbf{SAM-DT Pretraining (Dense Supervision):} During the high-fidelity SAM pretraining, we simulate partial observability at the input level by withholding the initial states of uninstrumented nodes, forcing the network to rely on the TGMI. However, because the SAM simulation provides ground-truth data for all physical volumes, we intentionally bypass the mask in the loss computation (setting $m_i=1$ for all $i \in \mathcal{V}_p$). This dense supervision explicitly penalizes predictions at both observed and uninstrumented locations, forcing the GNN-ODE to learn the true underlying physics required to correct TGMI initializations and reconstruct missing internal variables.

\textbf{Experimental Fine-Tuning (Sparse Supervision):} Conversely, during the sim-to-real fine-tuning phase on the physical facility data, ground-truth measurements for the internal volumes (e.g., heat exchangers and the mixing chamber) are physically unavailable. In this regime, the exact observability mask is enforced ($m_i=0$ for uninstrumented nodes), meaning these hidden states do not directly contribute to the computed loss. However, because the continuous-time latent dynamics are tightly coupled via the graph topology, these unmeasured nodes are implicitly updated. The gradients computed at the observable boundary nodes backpropagate through the spatial message-passing layers, allowing the hidden nodes to indirectly learn and adapt their thermal trajectories to the physical testbed while maintaining the energy-balance constraints established during pretraining.

\section{Case Study} \label{sec:case_study}

The development of the proposed GNN-ODE surrogate model is anchored on a control-oriented digital twin that serves as the primary environment for data generation, algorithm training, and future hardware-in-the-loop deployment. The system level thermal--hydraulic DT is implemented in the System Analysis Module (SAM)~\cite{hu2021samguide, hu2021sammanual} and mirrors a versatile thermal-fluid experimental testbed~\cite{lim2025ai, lim2026autonomous} designed for multipurpose cyber-physical research. Its thermal-hydraulic behavior is scaled to emulate SMR-relevant transients with emphasis on fluoride-salt-cooled concepts such as the Kairos Power FHR (KP-FHR)~\cite{blandford2020kairos}. Throughout this and next sections, we refer to this simulation environment as the \emph{SAM-based Digital Twin} (SAM-DT), which is used to generate synthetic trajectories (including ``virtual sensors'' at uninstrumented locations) and to provide a controllable, fully observable training environment prior to fine-tuning on experimental measurements.

\subsection{Experimental Facility and Test Framework}
\label{subsec: Exp facility}

\begin{figure}[htbp]
    \centering
    \includegraphics[width=0.65 \textwidth]{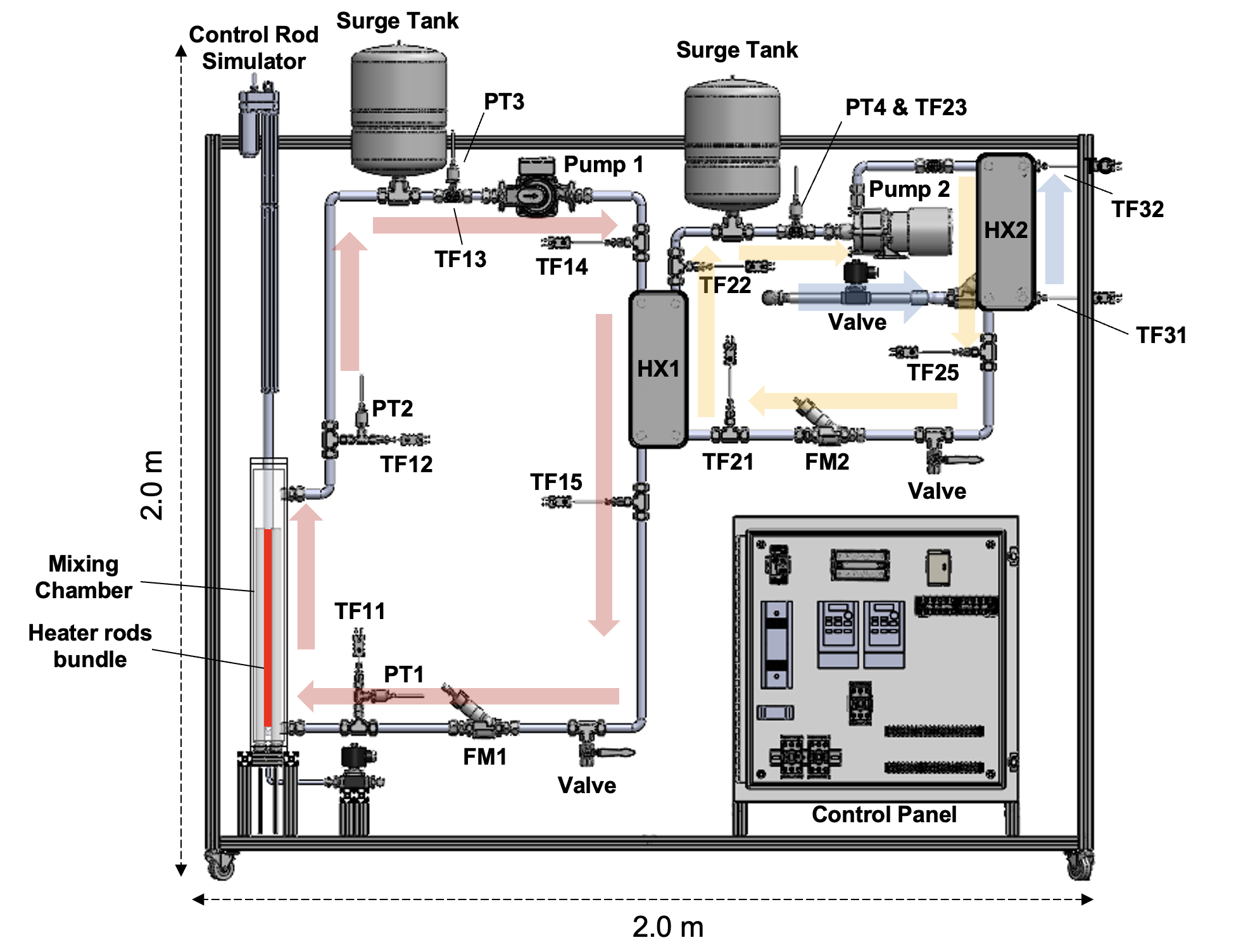}
    \caption{CAD model of the experimental thermal--hydraulic facility illustrating the three-loop configuration and the locations of key instrumentation and control components. Flow directions are indicated by colored arrows: primary loop (red), secondary loop (yellow), and tertiary loop (blue).}
    \label{fig:facility}
\end{figure}

The physical testbed (Figure~\ref{fig:facility}) comprises three hydraulically independent loops (primary, secondary, and tertiary) coupled through compact heat exchangers to transfer heat while operating near atmospheric pressure. Importantly, the facility is purely thermal-hydraulic: heat is supplied electrically, and the control-rod surrogate provides a \emph{reactivity-like} actuation channel for power--flow maneuvers without modeling reactor kinetics.

The primary loop includes a transparent heating chamber with four cartridge heaters in a \(2\times2\) lattice (\(16~\mathrm{kW}\) total), individually gated to command uniform or asymmetric power profiles, and a control-rod surrogate driven by a precision linear actuator (travel up to \(609.6~\mathrm{mm}\)). Heat is rejected sequentially through brazed-plate exchangers, and flow is regulated with variable-speed pumps in the primary/secondary loops; the \(19~\mathrm{mm}\) (3/4-inch) insulated stainless-steel piping minimizes heat losses.

In the experimental facility, state estimation and surrogate training rely on 29 logged signals, supported by a comprehensive instrumentation suite for model calibration and controller training. The thermal power delivered by the heater is determined by the product of the applied voltage and the resulting current. The hardware measurement and actuation uncertainties for these components are summarized in Table~\ref{tab:hardware_uncertainties}.

\begin{table}[hbt!]
    \centering
    \caption{Summary of experimental hardware measurement and actuation uncertainties.}
    \label{tab:hardware_uncertainties}
    \small
    \setlength{\tabcolsep}{3pt}
    \begin{tabularx}{\columnwidth}{@{}>{\raggedright\arraybackslash}X>{\raggedright\arraybackslash}X>{\centering\arraybackslash}p{0.22\columnwidth}@{}}
        \toprule
        \textbf{Component / Sensor} & \textbf{Measurement Type} & \textbf{Uncertainty} \\
        \midrule
        T-Type Thermocouples & Loop Temperatures & $\pm 1.0~^{\circ}\mathrm{C}$ \\
        J-Type Thermocouples & Heater Cartridge Temperatures & $\pm 2.2~^{\circ}\mathrm{C}$ \\
        Variable-Area Transmitters & Loop Flow Rates (up to $0.15~\mathrm{kg/s}$) & $\pm 5\%$ \\
        Gauge Pressure Transducers & System Pressures (up to $206.8~\mathrm{kPa}$) & $\pm 1\%$ \\
        Heater Controller & Actuated Thermal Power & $\pm 5\%$ \\
        \bottomrule
    \end{tabularx}
\end{table}

The measurement set includes 12 bulk coolant temperatures across the primary (TF11--TF15), secondary (TF21--TF25), and tertiary (TF31--TF32) loops; heater-rod temperatures (TH1--TH4), represented in this work by their average ($\mathrm{T_{H,avg}}$); pressures (PT1--PT4); loop flow rates (FM1--FM3); and electrical/thermal power. Main actuator commands (heater power, pump speeds, tertiary-loop valve position, and control-rod position) are logged as issued by the local control system to track the demand setpoint.
All field devices are supervised by an industrial PLC with interlocks, while an OPC-UA backbone exposes synchronized sensor streams and actuator set-points for real-time coupling with the SAM-based digital twin.

\begin{figure}[htbp]
\centering
\resizebox{0.85\textwidth}{!}{%
\begin{tikzpicture}[
    obs_node/.style={circle, draw=black, thick, fill=white, minimum size=1.2cm, inner sep=1pt, align=center, font=\small\bfseries},
    hid_node/.style={circle, draw=black, thick, dashed, fill=gray!15, minimum size=1.2cm, inner sep=1pt, align=center, font=\small\bfseries},
    act_node/.style={rectangle, rounded corners, draw=black, thick, fill=yellow!20, minimum height=0.8cm, minimum width=1.5cm, align=center, font=\small\bfseries},
    adv_L1/.style={-Stealth, line width=1.5pt, color=blue!80!black, shorten >=2pt, shorten <=2pt},
    adv_L2/.style={-Stealth, line width=1.5pt, color=orange!90!black, shorten >=2pt, shorten <=2pt},
    adv_L3/.style={-Stealth, line width=1.5pt, color=green!70!black, shorten >=2pt, shorten <=2pt},
    cond_edge/.style={Stealth-Stealth, line width=2pt, color=red!70, dashed, shorten >=2pt, shorten <=2pt},
    bound_edge/.style={-Stealth, line width=1.5pt, color=black, dotted, shorten >=2pt}
]

    \begin{scope}[rotate=90]

    \node[hid_node] (HX1_1) at (0.0, 5.6)   {$\mathrm{HX1\_L1}$};
    \node[obs_node] (TF15)  at (-1.8, 5.6)  {TF15};
    \node[obs_node] (TF11)  at (-1.8, 7.8)  {TF11};
    \node[hid_node] (Ch)    at (-1.8, 10.0) {$T_{ch}$};
    \node[obs_node] (TF12)  at (1.8, 10.0)  {TF12};
    \node[obs_node] (TF13)  at (1.8, 7.8)   {TF13};
    \node[obs_node] (TF14)  at (1.8, 5.6)   {TF14};

    \node[obs_node] (TH)    at (-4.6, 10.0) {$\mathrm{T_{H,avg}}$};
    \node[act_node] (PWR)   at ($(TH)+(0,-2.0)$) {Power};

    \node[align=center, font=\large\bfseries, color=blue!80!black] at (0, 7.8) {Loop 1 \\ (Primary)};

    \draw[adv_L1] (Ch)    -- (TF12);
    \draw[adv_L1] (TF12)  -- (TF13);
    \draw[adv_L1] (TF13)  -- (TF14);
    \draw[adv_L1] (TF14)  -- (HX1_1);
    \draw[adv_L1] (HX1_1) -- (TF15);
    \draw[adv_L1] (TF15)  -- (TF11);
    \draw[adv_L1] (TF11)  -- (Ch);

    \draw[bound_edge] (PWR) -- (TH);
    \draw[cond_edge]  (TH)  -- (Ch) node[midway, right, font=\footnotesize, color=black] {Convection};

    \node[hid_node] (HX1_2) at (0.0, 2.9)   {$\mathrm{HX1\_L2}$};
    \node[obs_node] (TF22)  at (1.8, 2.9)   {TF22};
    \node[obs_node] (TF23)  at (1.8, 0.7)   {TF23};
    \node[obs_node] (TF24)  at (1.8, -1.5)  {TF24};
    \node[hid_node] (HX2_1) at (0.0, -1.5)  {$\mathrm{HX2\_L2}$};
    \node[obs_node] (TF25)  at (-1.8, -1.5) {TF25};
    \node[obs_node] (TF21)  at (-1.8, 2.9)  {TF21};

    \node[align=center, font=\large\bfseries, color=orange!90!black] at (0, 0.7) {Loop 2 \\ (Secondary)};

    \draw[adv_L2] (HX1_2) -- (TF22);
    \draw[adv_L2] (TF22)  -- (TF23);
    \draw[adv_L2] (TF23)  -- (TF24);
    \draw[adv_L2] (TF24)  -- (HX2_1);
    \draw[adv_L2] (HX2_1) -- (TF25);
    \draw[adv_L2] (TF25)  -- (TF21);
    \draw[adv_L2] (TF21)  -- (HX1_2);

    \draw[cond_edge] (HX1_1) -- (HX1_2) node[midway, above, font=\footnotesize, color=black] {HX1};

    \node[hid_node] (HX2_2) at (0, -4.2)     {$\mathrm{HX2\_L3}$};
    \node[obs_node, fill=yellow!20] (TF31)  at (-2.5, -4.2)  {TF31 \\ (Inlet)};
    \node[obs_node] (TF32)  at (2.5, -4.2)   {TF32 \\ (Outlet)};

    \node[align=center, font=\large\bfseries, color=green!70!black] at (0, -6) {Loop 3 \\ (Tertiary)};

    \draw[adv_L3] (TF31)  -- (HX2_2);
    \draw[adv_L3] (HX2_2) -- (TF32);

    \draw[cond_edge] (HX2_1) -- (HX2_2) node[midway, above, font=\footnotesize, color=black] {HX2};

    \end{scope}

    \begin{scope}[shift={(-6, -7.5)}]
        \draw[thick, rounded corners=5pt, fill=white] (0,0) rectangle (10, 3.5);
        \node[font=\bfseries] at (5, 3.1) {Graph Topology Legend};
        
        \node[obs_node, minimum size=0.6cm] at (0.8, 2.3) {};
        \node[anchor=west, font=\small] at (1.2, 2.3) {Observable ($m_i = 1$)};
        
        \node[hid_node, minimum size=0.6cm] at (0.8, 1.4) {};
        \node[anchor=west, font=\small] at (1.2, 1.4) {Uninstrumented ($m_i = 0$)};

        \draw[cond_edge] (0.1, 0.5) -- (1.4, 0.5);
        \node[anchor=west, font=\small] at (1.4, 0.5) {Transverse exchange};

        \draw[adv_L1] (5.4, 2.3) -- (6.2, 2.3);
        \node[anchor=west, font=\small] at (6.2, 2.3) {Loop 1 Flow ($FM_1$)};

        \draw[adv_L2] (5.4, 1.4) -- (6.2, 1.4);
        \node[anchor=west, font=\small] at (6.2, 1.4) {Loop 2 Flow ($FM_2$)};
        
        \draw[adv_L3] (5.4, 0.5) -- (6.2, 0.5);
        \node[anchor=west, font=\small] at (6.2, 0.5) {Loop 3 Flow ($FM_3$)};
    \end{scope}

\end{tikzpicture}%
}
\caption{Horizontal, compact representation of the 3-loop thermal system graph topology.}
\label{fig:vertical_topology}
\end{figure}
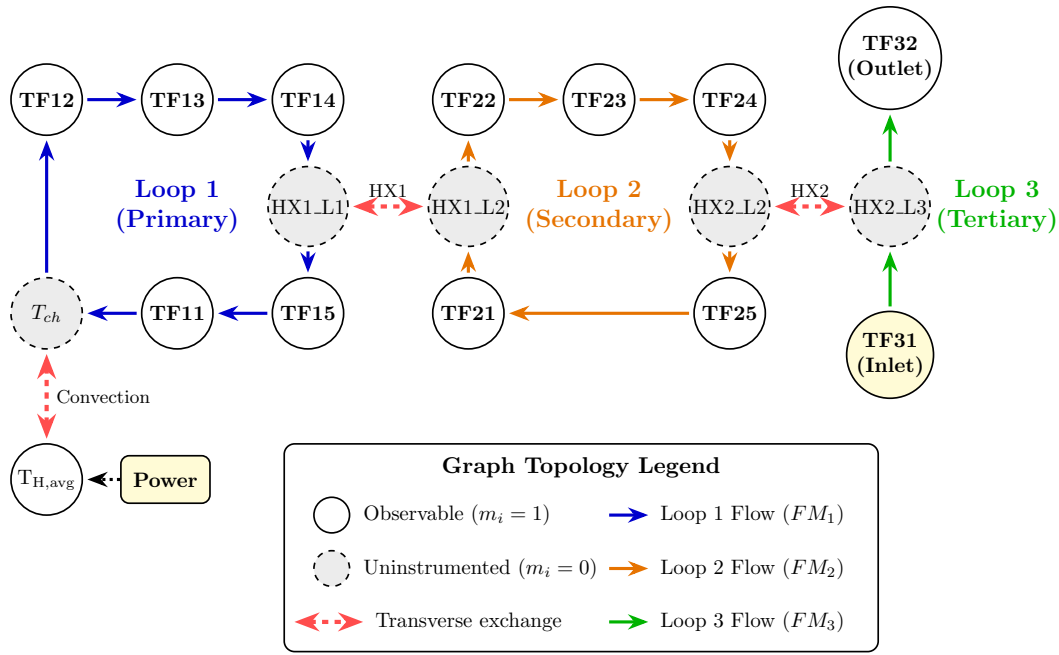

To support the partial-observability training strategy described in Section~\ref{sec:spatial_modeling}, we define a consistent mapping between facility channels and graph nodes that is shared by both experiments and SAM. The plant is discretized into 17 control volumes, logically structured as a directed graph representing the physical thermal-hydraulic connectivity. Figure~\ref{fig:vertical_topology} illustrates this compact vertical graph topology, highlighting the flow of advection and transverse heat exchange across the three loops. Point-wise thermocouple (TF) measurements are assigned to the corresponding volumes to reconcile experimental readings with volume-averaged simulation states, as summarized in Table~\ref{tab:node_mapping}. As shown in the table and the accompanying topology figure, this case study includes 12 observable nodes ($m_i=1$) and five uninstrumented internal volumes (the mixing chamber and heat-exchanger fluid nodes, $m_i=0$).

\begin{table}[t]
\centering
\caption{Nodal discretization and signal mapping scheme (``Uninstrumented'' nodes are latent states with $m_i=0$).}
\label{tab:node_mapping}
\footnotesize
\setlength{\tabcolsep}{3pt}
\renewcommand{\arraystretch}{1.05}
\begin{tabularx}{\columnwidth}{@{} c l c X @{}}
\toprule
\textbf{Idx} & \textbf{Sensor} & \textbf{Mask} & \textbf{GNN--ODE node name}\\
\midrule
0  & $\mathrm{T_{H,avg}}$      & $m_i=1$ & Heater rods average temperature \\
1  & $\mathrm{T_{ch}}$   & $m_i=0$ & Mixing chamber average temperature \\
2  & TF12              & $m_i=1$ & Leg 1 (hot), segment 1 \\
3  & TF13              & $m_i=1$ & Leg 1 (hot), segment 2 \\
4  & TF14              & $m_i=1$ & Leg 1 (hot), segment 3 \\
5  & HX1\_L1    & $m_i=0$ & Heat Exchanger 1, primary side \\
6  & TF15              & $m_i=1$ & Leg 1 (cold), segment 1 \\
7  & TF11              & $m_i=1$ & Leg 1 (cold), segment 2 \\
8  & HX1\_L2    & $m_i=0$ & Heat Exchanger 1, secondary side \\
9  & TF22              & $m_i=1$ & Leg 2 (hot), segment 1 \\
10 & TF23              & $m_i=1$ & Leg 2 (hot), segment 2 \\
11 & TF24              & $m_i=1$ & Leg 2 (hot), segment 3 \\
12 & HX2\_L2    & $m_i=0$ & Heat Exchanger 2, primary side \\
13 & TF25              & $m_i=1$ & Leg 2 (cold), segment 1 \\
14 & TF21              & $m_i=1$ & Leg 2 (cold), segment 2 \\
15 & HX2\_L3    & $m_i=0$ & Heat Exchanger 2, secondary side \\
16 & TF31              & $m_i=1$ & Loop 3 inlet ($T_{c,in}$) \\
17 & TF32              & $m_i=1$ & Loop 3 outlet ($T_{out}$) \\
\bottomrule
\end{tabularx}
\end{table}

\subsection{SAM-based digital-twin dataset generation}
\label{subsec: SAM-DT}
\begin{figure}[htbp]
    \centering
    \includegraphics[width=0.75\textwidth]{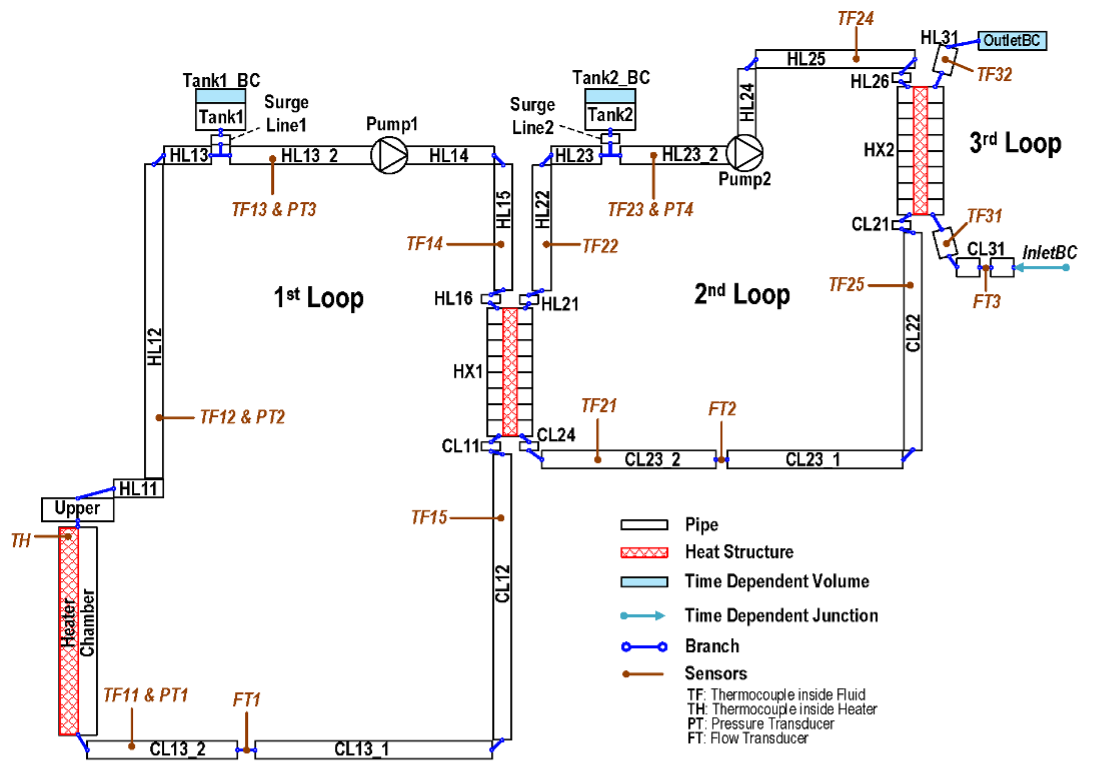}
    \caption{SAM-based digital twin model of the experimental thermal-fluid facility used for synthetic trajectory generation.}
    \label{fig:sam_model}
\end{figure}

We combine measured trajectories from the thermal-fluid testbed with synthetic trajectories generated by the SAM-DT, which enables data generation beyond experimental operating limits. Figure~\ref{fig:sam_model} illustrates the implemented SAM-DT model used to produce the synthetic training trajectories. 

Crucially, while the SAM model captures the foundational thermodynamic structure of the facility, it is an idealized 1D approximation that does not precisely replicate the true experimental behavior. Specifically, across a broad range of operations, a growing divergence between the SAM predictions and experimental measurements is observed at high power values. This discrepancy stems from complex, unmodeled 3D mixing phenomena in the chamber and nonlinear parasitic heat losses to the ambient environment that scale dynamically with elevated temperatures. Consequently, the synthetic dataset acts not as a perfect surrogate for reality, but rather as a rigorous pretraining environment. It broadens the training distribution and provides ``virtual sensors'' at uninstrumented locations (as detailed in Section~\ref{sec:training_strategy}), establishing foundational physical priors that govern the topology-guided state reconstructions.

Synthetic trajectories are generated via a parameter sweep in which initial conditions and actuator settings are sampled using Latin Hypercube Sampling (LHS; seed 42; 1000 designs) and augmented with 50 edge-case designs. The sweep spans loop flow operating points (Loops 1--2 valve-controlled and Loop 3 directly prescribed), heater power, and initial wall temperatures, while enforcing constraints such as a 16~kW power cap and reduced power limits under low-flow conditions.

To excite transient behavior, each simulation includes two randomly instantiated perturbations within the active window (start times 50--200~s), with a 10~s ramp duration and a minimum 20~s separation. Perturbations act on heater power and/or individual loop flows, yielding diverse single- and paired-input transients. The deliberate 20~s separation is enforced because simultaneous control actions are strictly prohibited in both our testbed and commercial nuclear plants. By adhering to this ``one control at a time'' human-performance standard \cite{nrc_ml102120052, doe_hdbk_1028_2009}, this scenario represents an operationally realistic, yet conservatively challenging, test. Across the sweep we generated 700 time-series CSV files, each containing 300~s of active simulation preceded by a 200~s initialization period. 

Finally, SAM employs adaptive time stepping for robustness during strong perturbations (down to $5\times 10^{-4}$~s in challenging cases). Otherwise, the nominal step is 0.5~s and results are recorded at 1~s resolution (every two solver steps), producing an effective sampling interval spanning roughly $10^{-3}$ to 1~s. This variable-rate sampling is naturally handled by the continuous-time Neural ODE formulation.

More broadly, the continuous-time formulation naturally handles heterogeneous sensing rates: when a sensor is unavailable at a given instant, the corresponding mask is set to $m_i=0$ and the loss is evaluated only at observed nodes, avoiding ad hoc interpolation assumptions.

To evaluate model performance under representative transient operating conditions, we assessed four held-out scenarios specifically designed to test generalization and predictive accuracy. These scenarios include time-varying control actions with coupled perturbations in power and loop flow rates, as summarized in Table~\ref{tab:transient_scenarios}.

\begin{table}[hbt!]
  \caption{Transient scenarios used in this study. Each scenario is 300~s. long and is being generated by SAM model.}
  \label{tab:transient_scenarios}
  \small
  \setlength\tabcolsep{4pt}
  \begin{tabularx}{\columnwidth}{@{}lX@{}}
    \toprule
    \textbf{Scenario} & \textbf{Description} \\
    \midrule
    Stepwise heat-load ramp & Heat load is increased from 1$\rightarrow$5\,kW in 1\,kW increments every 60\,s, with all three loop flow rates fixed at 0.14\,kg/s. \\
    Cascaded loop-flow changes & Cascaded flow-rate changes at 7.5\,kW: Loop~1, 0.14$\rightarrow$0.1\,kg/s at $t=30$\,s; Loop~2, 0.1$\rightarrow$0.12\,kg/s at $t=40$\,s; and Loop~3, 0.07$\rightarrow$0.14\,kg/s at $t=50$\,s. \\
    \bottomrule
  \end{tabularx}
\end{table}

\subsection{Experimental Fine-Tuning and Sim-to-Real Transfer}
\label{subsec:fine_tuning}

To evaluate the sim-to-real transferability of the learned physical priors, we deployed the SAM-trained GNN-ODE surrogate on the experimental facility data. As established in Section~\ref{subsec: SAM-DT}, the physical testbed exhibits fundamental thermodynamic differences from the idealized 1D SAM simulation, including complex 3D mixing phenomena, nonlinear parasitic heat losses, and distinct thermal inertias. 

To bridge this sim-to-real gap, the SAM-trained surrogate underwent a rapid fine-tuning phase using a strictly limited experimental dataset. The raw experimental logs were preprocessed and partitioned into discrete 5-minute transient sequences, yielding a sparse dataset of exactly 30 sequences for training, 9 for validation, and 1 held-out test case (evaluated subsequently in the Results section). The ability to successfully adapt the continuous-time model using very little computational time and minimal physical data serves as a compelling proof-of-concept. It strongly indicates that future topology transfer studies, in which the model is transferred to similar physical facilities, are highly feasible.

However, fine-tuning a highly parameterized model on such a small dataset introduces the risk of overfitting and loss of simulation-derived priors. Applying a uniform, high learning rate would cause the model to rapidly overfit to the limited experimental training sequences, effectively memorizing specific temperature trajectories rather than generalized physics. 

To prevent this, we applied a targeted fine-tuning phase utilizing Discriminative Learning Rates, inspired by Layer-wise Learning Rate Decay (LLRD) \cite{howard2018universallanguagemodelfinetuning}. All groups were decayed jointly via a cosine annealing schedule over the full 500-epoch training, preserving the relative adaptation ratios throughout, but were assigned specific rates based on their required physical adaptation:
 
\begin{itemize}
    \item \textbf{GNN Layers ($5 \times 10^{-5}$):} The lowest learning rate was applied to the spatial message-passing layers. Because the macroscopic facility topology (e.g., advective piping routes and heat exchanger connections) remains identical between SAM and reality, these weights require minimal adjustment. Moving them too aggressively would destroy the valid spatial heat-transfer priors.
    \item \textbf{Actuator Projection ($1 \times 10^{-4}$):} A moderate learning rate was assigned to the input projection mapping. While SAM operates on exact physical power delivery (Watts), the experimental dataset relies on the commanded power setpoint. The general relationship remains physically proportional, but the exact mapping into the latent space requires moderate adaptation.
    \item \textbf{Neural ODE and Output Head ($5 \times 10^{-4}$):} The fastest learning rate was applied to the continuous-time dynamics engine and the decoder. Because the SAM simulation inherently differs from reality regarding insulation efficiency, environmental heat losses, and sensor response times, the ODE solver requires maximum flexibility to learn the true physical thermal inertias and temporal time constants of the testbed.
\end{itemize}

Initial experiments with a fully frozen encoder confirmed this risk: the model achieved low training loss under teacher forcing but failed to produce stable autonomous rollouts, with validation loss diverging to 10$\times$ the training value.

\section{Results}
\label{sec:results}

We report performance primarily using Mean Absolute Error (MAE) in Kelvin, as it provides a direct and physically interpretable measure of thermal prediction error across the facility operating range. Relative errors are less informative in this setting: percentage normalization can obscure practically relevant absolute deviations in high-baseline Kelvin signals and can become unstable when referenced to small local temperature differences. Over the observed operating intervals (non-heater channels: 280.07--362.69 K; heater channel $T_{H,\mathrm{avg}}$: 293.24--411.24 K), MAE provides clearer engineering interpretability for rollout-quality assessment.

\subsection{Topology-Guided Missing-nodes Initializer Performance}

We first assess TGMI performance for reconstructing temperatures at uninstrumented nodes. Table~\ref{tab:missing-node-recon} reports reconstruction accuracy with and without the global context vector $\tilde{\mathbf{u}}$, as defined in Eq.~\ref{eq:tgmi}.

Including the global context vector substantially improves reconstruction quality, reducing MAE by up to 2~K. The stronger dependence of HX2\_L3 on $\tilde{\mathbf{u}}$ likely reflects its sparser instrumented neighborhood relative to the other missing nodes. By contrast, the chamber node is already strongly informed by global dynamics through its adjacency to the heater node, which has incoming power input and whose transverse-edge coupling is highly flow dependent.

\begin{table}[H]
\centering
\caption{Reconstruction performance for missing nodes on validation dataset.}
\label{tab:missing-node-recon}
\begin{tabular}{lccccc}
\toprule
Node & $R^2$ (w/o $\tilde{\mathbf{u}}$) & MAE (w/o $\tilde{\mathbf{u}})$ & $R^2$ & MAE & $\Delta$ MAE \\
\midrule
Ch      & 0.99989 & 0.076 & 0.99996 & 0.064 & -0.012 \\
HX1\_L1 & 0.93401 & 2.322 & 0.99328 & 0.686 & -1.637 \\
HX1\_L2 & 0.92213 & 1.980 & 0.99537 & 0.448 & -1.532 \\
HX2\_L2 & 0.93471 & 1.911 & 0.98953 & 0.632 & -1.279 \\
HX2\_L3  & 0.88419 & 2.559 & 0.99196 & 0.559 & -2.000 \\
\bottomrule
\end{tabular}
\end{table}
              
\subsection{Forecasting performance}
\label{subsec: forecasting}

During inference, forecasting is performed as a fully autoregressive rollout. At \(t_0\), the 5 uninstrumented node temperatures are initialized once using TGMI, while the 12 instrumented nodes are taken from measurements, forming the complete 17-node initial state. From that point onward, the model advances the full state recursively with no further TGMI updates and no teacher forcing.

With the velocity channel enabled, the velocity input at step \(t\) is computed from successive predicted states,
\begin{equation}
    \widehat{\dot{\mathbf T}}_t=\frac{\hat{\mathbf T}_t-\hat{\mathbf T}_{t-1}}{\Delta t_{t-1}},
\end{equation}
and the Neural ODE is integrated using the actual per-step \(\Delta t_t\). Thus, after one-time TGMI initialization at \(t_0\), the trajectory is generated entirely by autoregressive GNN-ODE dynamics.

To quantify inference speed, we benchmarked no-plot rollout performance on 10 SAM trajectories, each 300~s long. On an NVIDIA GeForce RTX~4090, the mean wall-clock rollout time was 2.861~s, corresponding to a speedup factor of 104.49 relative to simulated time (9.544~ms per simulated second).

This inference speed makes parallel ensemble rollouts practical for uncertainty propagation. The propagated uncertainty levels are based on the experimental instrumentation uncertainties described in Subsection~\ref{subsec: Exp facility}.

To further characterize scalability, we benchmarked no-plot rollouts for a single model ($M=1$) and ensembles with $M=16,32,64$. Table~\ref{tab:ensemble_rollout_perf} summarizes the resulting speedup factors for both GPU and CPU execution (AMD Ryzen Threadripper PRO 5995WX, 64 cores, 128 threads).

\begin{table}[t]
\centering
\caption{No-plot rollout speedup factors for single-model and ensemble inference, computed as simulated (predicted) time divided by wall-clock runtime.}
\label{tab:ensemble_rollout_perf}
\small
\setlength\tabcolsep{3pt}
\begin{tabular}{@{}lcccc@{}}
\toprule
\textbf{Device/Mode} & $M=1$ & $M=16$ & $M=32$ & $M=64$ \\
\midrule
\shortstack[t]{CPU 1-thread} & 146.02 & 35.31 & 16.26 & 8.47 \\
\shortstack[t]{CPU 128-thread} & 63.27 & 24.52 & 19.41 & 14.50 \\
\shortstack[t]{GPU (RTX~4090)} & 104.49 & 66.43 & 52.07 & 36.57 \\
\bottomrule
\end{tabular}
\end{table}

These results indicate that CPU single-thread execution is optimal for $M=1$, whereas CPU multithreading becomes increasingly beneficial as ensemble size grows relative to CPU single-thread execution. A likely explanation is that per-step graph construction and small-kernel operations dominate runtime, making heavy threading and GPU kernel-launch overhead less advantageous. For medium-to-large ensembles ($M\geq16$), GPU execution provides the best overall performance, with the largest advantage observed at $M=64$.

\begin{figure}[!t]
    \centering
    \subfloat[Stepwise heat-load ramp\label{fig:s1}]{%
        \includegraphics[width=0.49\textwidth]{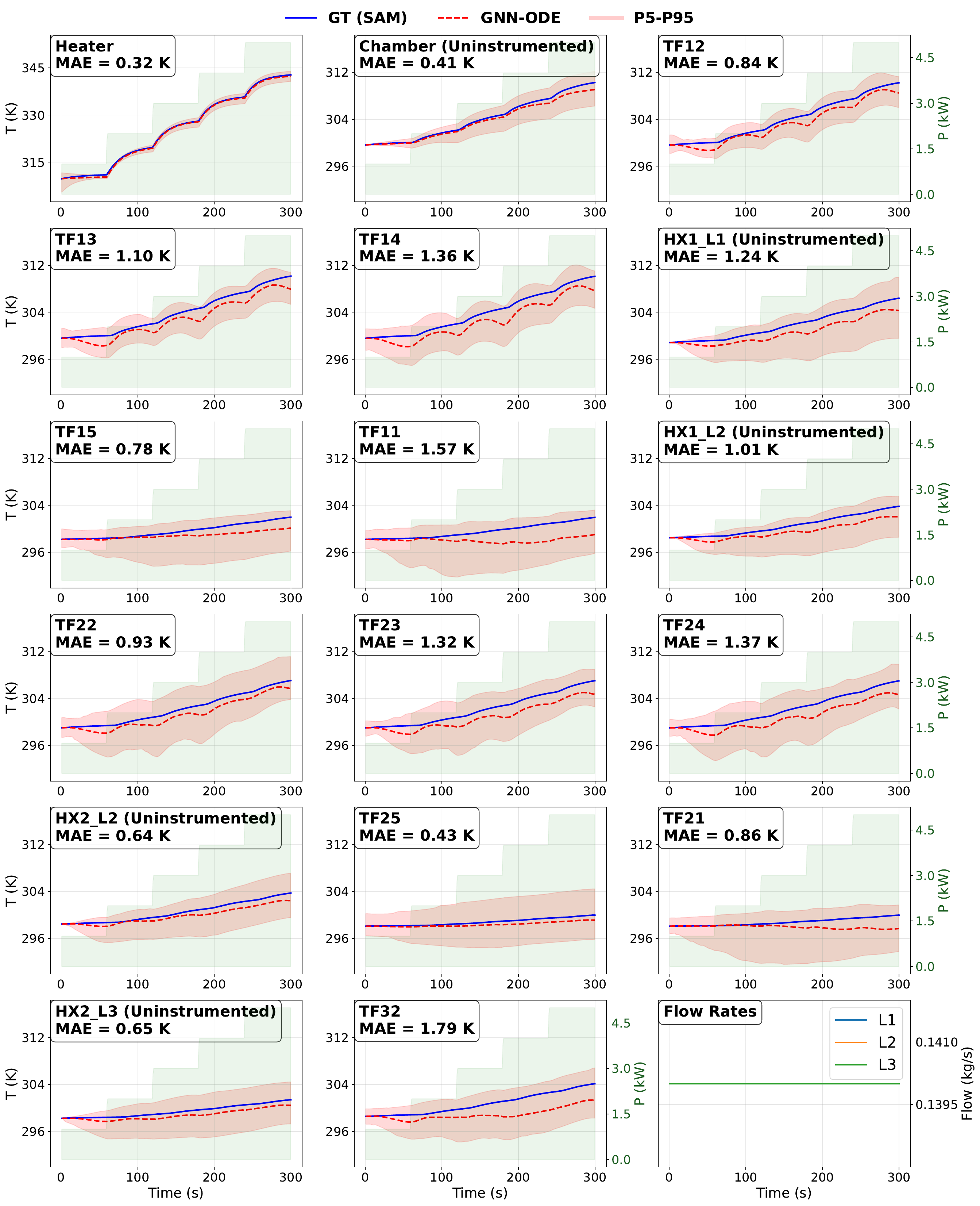}}
    \subfloat[Cascaded loop-flow changes\label{fig:s4}]{%
        \includegraphics[width=0.49\textwidth]{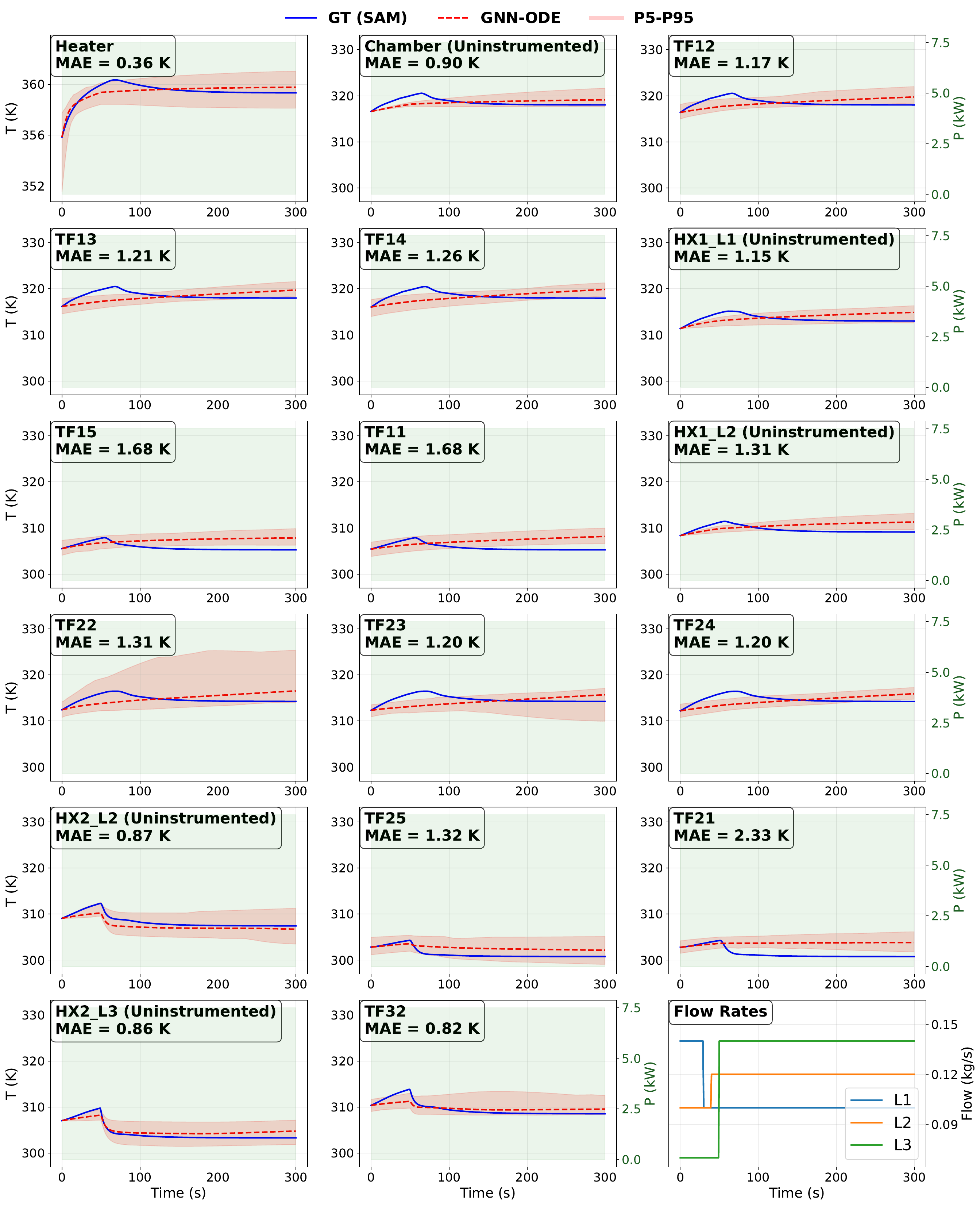}}
    \caption{Forecasting comparison across the two transient scenarios listed in Table~\ref{tab:transient_scenarios}. The red dashed line corresponds to non-perturbed GNN-ODE predictions, while the red background indicates the predictive uncertainty band between the 5th and 95th percentiles. The green background denotes the imposed power profile on the secondary y-axis. GT (SAM) denotes SAM ground truth, which is used to compare against GNN-ODE predictions and compute MAE.}
    \label{fig:scenario_rollouts}
\end{figure}

Performance of the GNN-ODE model for the four scenarios listed in Table~\ref{tab:transient_scenarios} applying the parallel ensemble rollouts for uncertainty propagation ($M=64$) are presented in Figs.~\ref{fig:s1}--\ref{fig:s4}.

The results show high predictive fidelity for the heater and heat-exchanger nodes, indicating that the model captures transverse coupling effectively. In contrast, errors accumulate primarily along streamwise advection pathways within the loops, although part of this drift is mitigated at downstream heat-exchanger locations. Scenario-level comparisons support this interpretation: profile quality degrades more noticeably in transients with rapid flow-rate variations, where advection-dominated dynamics are stronger.

\begin{figure}[h]
    \centering
    \subfloat[Observed plant nodes\label{fig:forecasting_observed}]{%
        \includegraphics[width=0.49\columnwidth]{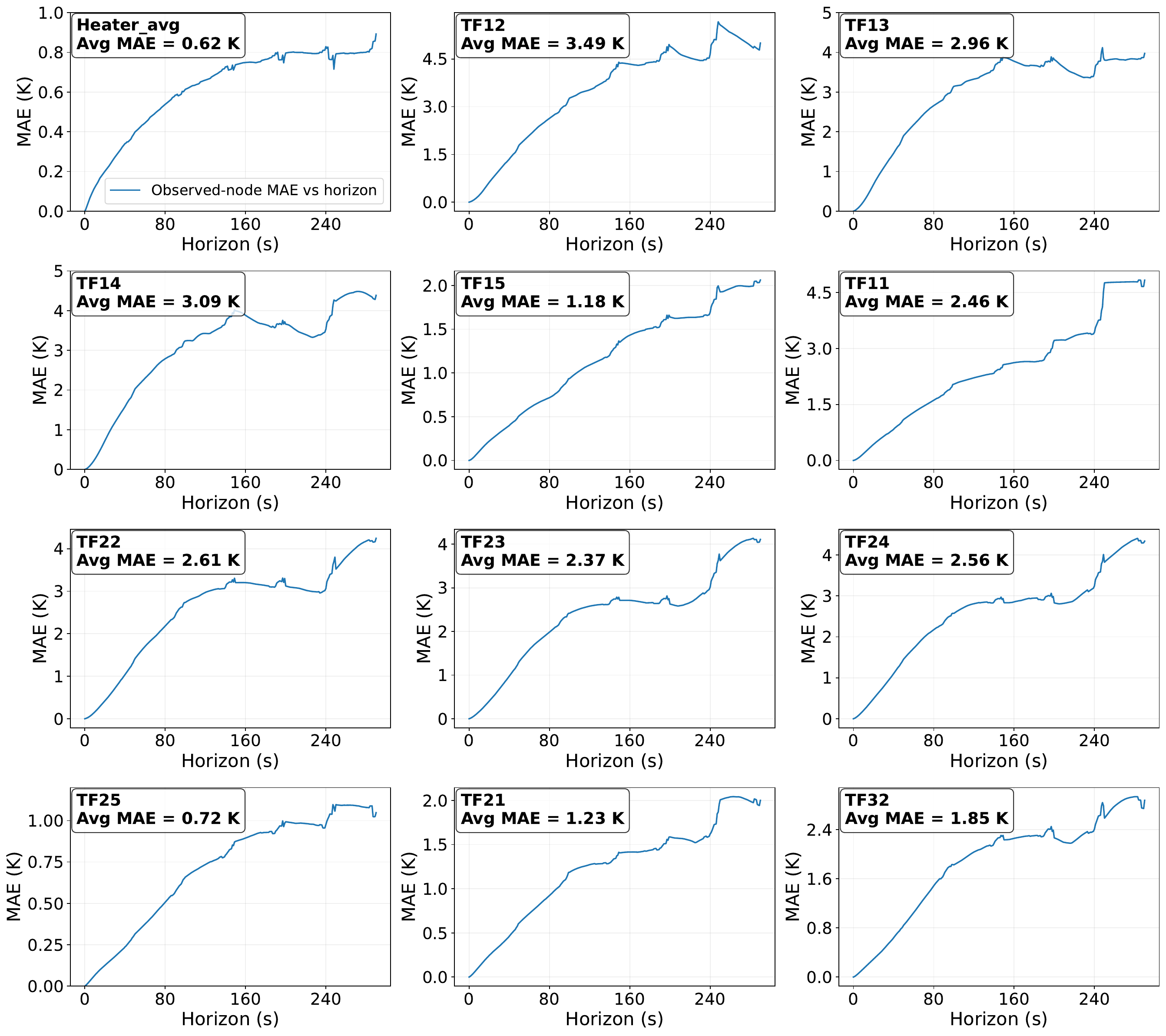}}
    \hfill        
    \subfloat[Unobserved (missing) plant nodes\label{fig:forecasting_missing}]{%
        \includegraphics[width=0.49\columnwidth]{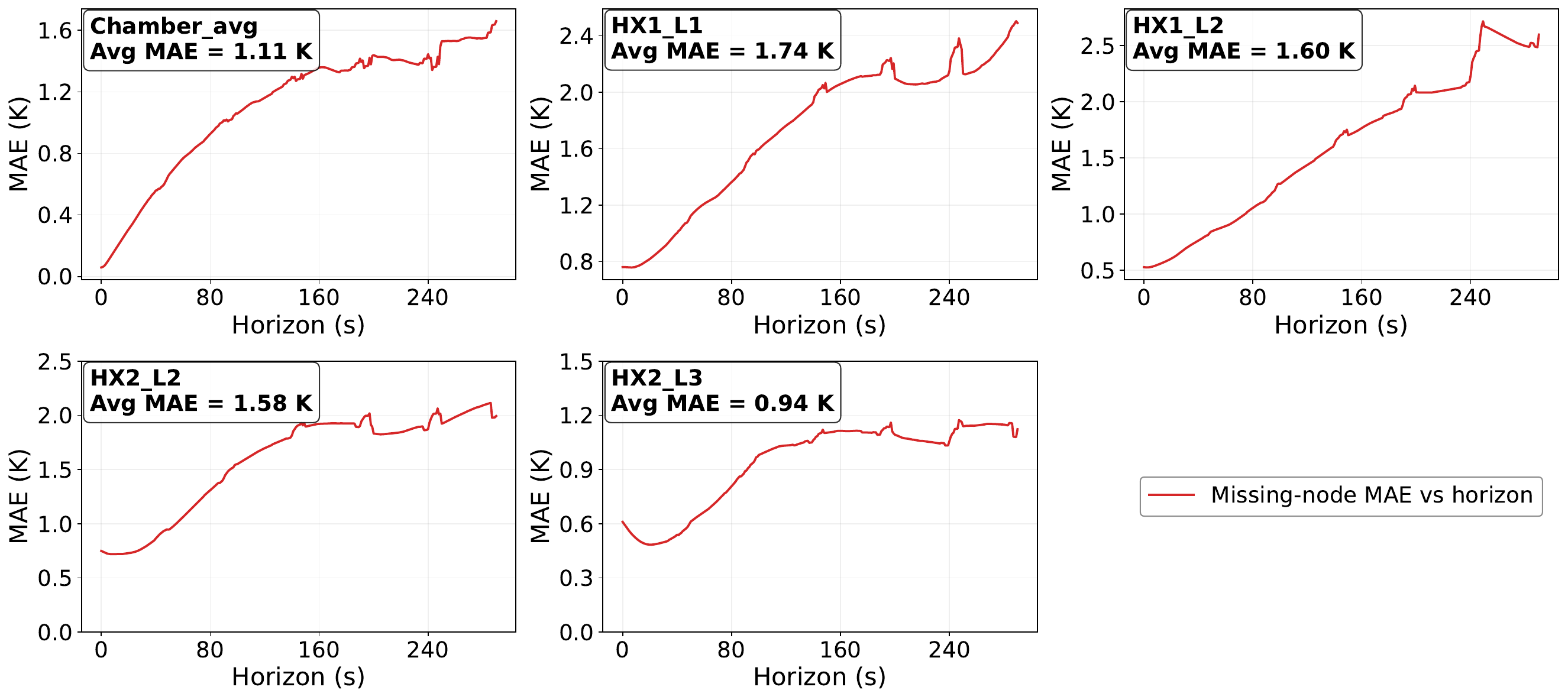}}
    \caption{Long-horizon rollout Mean Absolute Error (MAE) versus forecast horizon on held-out SAM test transients for (a) observed nodes and (b) unobserved (missing) nodes.}
    \label{fig:forecasting_horizon}
\end{figure}

Importantly, error growth is assessed over the whole testing data lengths of 300~s, whereas training was performed with unroll windows of 64 steps (approximately 60~s). The 64-step choice reflects a practical trade-off between computational cost and the marginal accuracy gains obtained from longer training sequences. All rollouts are performed on held-out SAM-generated test transients to assess generalization, including the challenging case of forecasting uninstrumented (missing) nodes. As expected, rollout error increases with prediction horizon for both missing and observed node subsets as can be seen in Figs.~\ref{fig:forecasting_missing}--\ref{fig:forecasting_observed} and Tables~\ref{tab:forecast-missing}--\ref{tab:forecast-observed}. Notably, for the missing second heat-exchanger nodes (HX2\_L2 and HX2\_L3), the error can initially decrease over the first \(\sim\)30~s as the surrogate corrects the linear-regression initialization described in Subsection~\ref{subsec:partial_obs} through autoregressive inference. 

Uncertainty analysis is performed using parallel ensemble rollouts ($M=64$), where perturbations were sampled solely from the experimentally characterized hardware uncertainties summarized in Table~\ref{tab:hardware_uncertainties}. Results indicate that the dominant contribution to forecast spread comes from uncertainty in control inputs, primarily heater power and loop flow rates, which directly drive system-level thermal trajectories. By contrast, temperature uncertainty has a secondary role and mainly enters through initialization of the state estimate at rollout start, after which input-driven effects dominate uncertainty propagation. 

\begin{table}[h]
\centering
\caption{Forecasting rollout summary for observed nodes (GNN-ODE forecast).}
\label{tab:forecast-observed}
\begin{tabular}{lcccccc}
\toprule
Node & 10s & 30s & 60s & 120s & 180s & 300s \\
\midrule
Heater\_avg & 0.14 & 0.44 & 0.76 & 1.52 & 2.31 & 0.55 \\
TF12 & 0.23 & 0.94 & 1.78 & 2.71 & 2.96 & 6.59 \\
TF13 & 0.26 & 1.11 & 2.03 & 2.83 & 2.83 & 4.68 \\
TF14 & 0.28 & 1.24 & 2.29 & 3.34 & 3.46 & 4.86 \\
TF15 & 0.10 & 0.31 & 0.58 & 0.95 & 1.32 & 2.53 \\
TF11 & 0.16 & 0.55 & 1.09 & 1.96 & 2.51 & 2.97 \\
TF22 & 0.16 & 0.62 & 1.34 & 2.13 & 2.55 & 5.27 \\
TF23 & 0.16 & 0.60 & 1.25 & 2.06 & 2.40 & 4.65 \\
TF24 & 0.20 & 0.73 & 1.49 & 2.27 & 2.62 & 5.11 \\
TF25 & 0.06 & 0.17 & 0.34 & 0.55 & 0.75 & 1.14 \\
TF21 & 0.11 & 0.32 & 0.61 & 0.99 & 1.24 & 1.49 \\
TF32 & 0.12 & 0.39 & 0.86 & 1.45 & 1.89 & 4.00 \\
AVERAGE & 0.17 & 0.62 & 1.20 & 1.90 & 2.24 & 3.66 \\
\bottomrule
\end{tabular}
\end{table}

\begin{table}[h]
\centering
\caption{Forecasting rollout summary for missing nodes (virtual sensors + GNN-ODE forecast).}
\label{tab:forecast-missing}
\begin{tabular}{lcccccc}
\toprule
Node & 10s & 30s & 60s & 120s & 180s & 300s \\
\midrule
Chamber\_avg & 0.18 & 0.49 & 0.85 & 1.34 & 1.87 & 1.17 \\
HX1\_L1 & 0.74 & 0.87 & 1.22 & 1.72 & 2.10 & 3.12 \\
HX1\_L2 & 0.53 & 0.65 & 0.87 & 1.29 & 1.53 & 2.82 \\
HX2\_L2 & 0.72 & 0.76 & 0.93 & 1.22 & 1.52 & 2.11 \\
HX2\_L3 & 0.52 & 0.51 & 0.66 & 0.88 & 1.06 & 1.71 \\
AVERAGE & 0.54 & 0.66 & 0.91 & 1.29 & 1.62 & 2.18 \\
\bottomrule
\end{tabular}
\end{table}

The forecasting accuracy achieved over a 60~s prediction horizon satisfies the requirements for our intended control-oriented deployments. This horizon offers a practical balance between fidelity and computational efficiency, and it motivates direct integration of the surrogate into future studies on advanced control design.  

Importantly, the model's generalization is driven by its ability to recover physically meaningful heat-transfer scaling without explicit supervision. Operating closest to the physical input representation, the first GNN layer learned an exponent of $\alpha \approx 0.87$ from Eq.~\ref{eq: HTC} for the heat-exchanger edges, within approximately 10\% of the Dittus-Boelter correlation's Reynolds number exponent (0.8) used in the SAM reference model for forced convection. However, for the heater-to-chamber coupling, the learned exponent was consistently lower ($\alpha \approx$ 0.69-0.76).

This divergence reflects a compound effect of multiple physical and modeling factors. First, it captures the spatial complexities of the rod-bundle geometry, thermal mixing, and volume-averaging enforced by the lumped GNN node. Second, it mathematically compensates for the surrogate's simplifying assumption of a uniform bulk flow rate. In reality—and in the SAM simulation—local flow velocity over the heating rods is heavily modified by natural circulation and buoyancy forces. 

Because the GNN is conditioned only on the macroscopic loop flow rate, it must implicitly account for these localized mixed-convection dynamics and geometric complexities. The separation between the heat exchangers and heater exponents shows that the network independently discovers these distinct flow sensitivities, even though both utilized $\alpha = 0.8$ correlations in the baseline SAM model. Overall, these findings demonstrate that the surrogate is not merely fitting trajectories, but is actively learning interpretable constitutive relationships to bridge the gap between simplified inputs and complex physical realities.

\subsection{Experimental Evaluation and Hidden State Inference}
\label{subsec:exp_results}

As detailed in Section~\ref{subsec:fine_tuning}, bridging the sim-to-real gap required adapting the SAM-trained surrogate to the physical testbed using a targeted, discriminative fine-tuning strategy. To evaluate the success of this transfer, we deploy the fine-tuned GNN-ODE model on the specifically held-out experimental test sequence.

A primary numerical difficulty encountered during training was inherent sensor noise. Because both the model's predictive mechanism and the loss formulation are driven by continuous-time state derivatives ($d\mathbf{T}/dt$), high-frequency fluctuations are intrinsically amplified during differentiation, so minor temperature perturbations can produce disproportionately large apparent rates of change and mask the underlying macroscopic thermal dynamics. To mitigate this effect, raw experimental temperature signals were pre-processed with a Savitzky--Golay filter \cite{savitzky1964smoothing, feng2010temperature} (window length 7, second-order polynomial), which attenuated high-frequency noise while preserving thermal transient shapes; this is critical because the training targets are finite-difference rates ($d\mathbf{T}/dt$). The filter reduced rate standard deviations by 40--50\% across most sensor nodes without introducing phase lag in the transient response.

\begin{figure}
    \centering
    \includegraphics[width=0.95\textwidth]{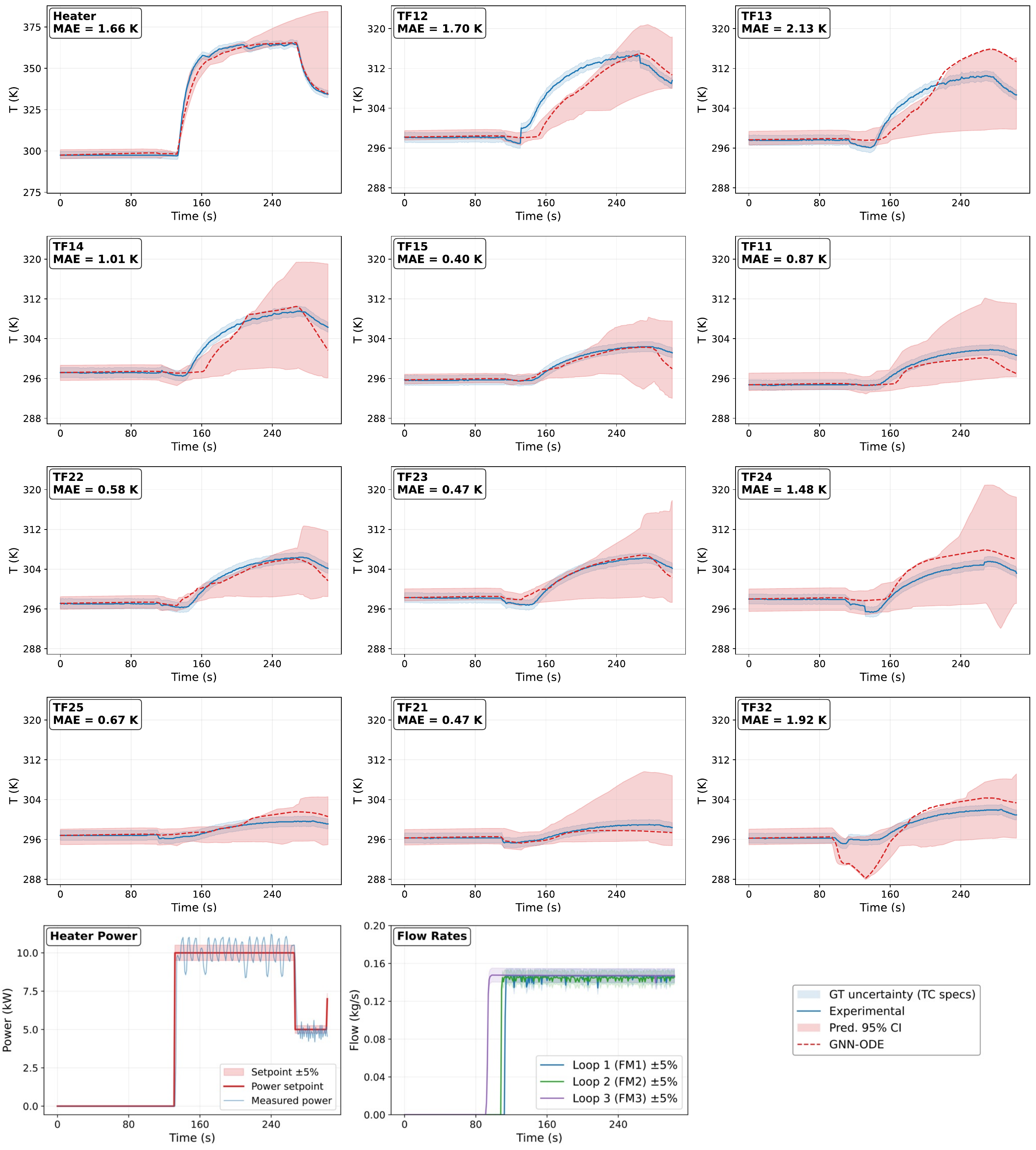}
    \caption{Surrogate predictions versus experimental measurements for observable facility nodes during steep power changes ($0 \rightarrow 10 \rightarrow 5 \rightarrow 7$~kW). Shaded bands denote 95th-percentile confidence intervals from $M=64$ parallel ensemble rollouts.}
    \label{fig:exp_observable_nodes}
\end{figure}

\begin{figure}
    \centering
    \includegraphics[width=0.95\textwidth]{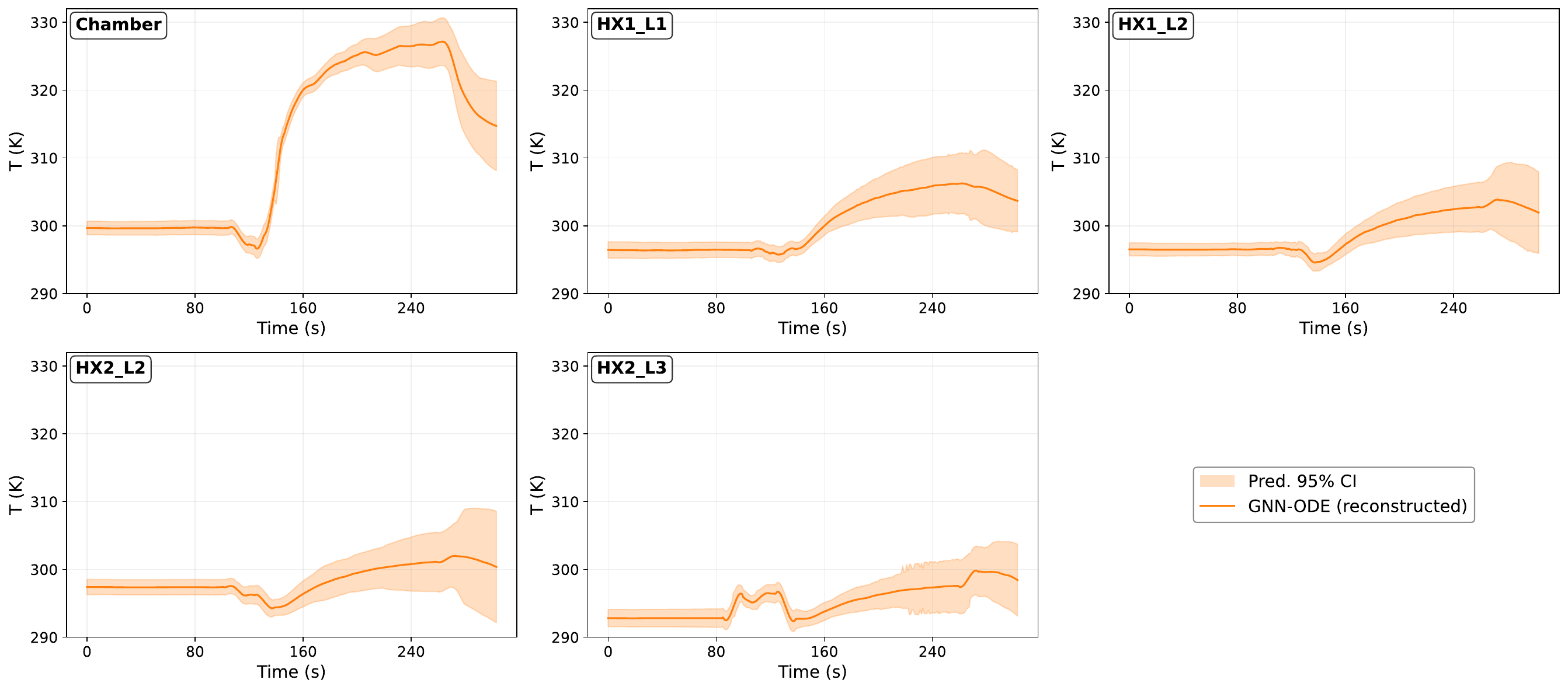}
    \caption{Inferred thermal trajectories for permanently uninstrumented (hidden) nodes during steep power changes ($0 \rightarrow 10 \rightarrow 5 \rightarrow 7$~kW). Initial states are seeded by TGMI; shaded bands show 95th-percentile confidence intervals from $M=64$ ensemble rollouts.}
    \label{fig:exp_missing_nodes}
\end{figure}

Despite these challenging real-world conditions, the surrogate remains highly predictive. To mitigate the destabilizing effect of derivative-amplified noise during evaluation, the heater-power input was prescribed using the commanded setpoint rather than the raw, fluctuating measured signal (both are shown in Figure~\ref{fig:exp_observable_nodes}). This smoother forcing input suppresses high-frequency disturbances that could otherwise destabilize the continuous-time ODE rollouts. Flow-rate noise was comparatively acceptable and therefore left unsmoothed; however, mild smoothing analogous to the temperature preprocessing, with a shorter window to preserve true pump and valve adjustments, could further improve robustness in future work.

We evaluated the model's performance on the held-out test case: a steep, multi-step experimental transient ($0 \rightarrow 10 \rightarrow 5 \rightarrow 7$~kW) executed while holding primary, secondary, and tertiary flow rates constant. Predictive uncertainty was propagated using the previously established ensemble protocol ($M=64$), and the resulting variance bands indicate that control-input variability remains the dominant driver of forecast uncertainty. This test case also clearly illustrates the uncertainty dynamics. As shown in Figure~\ref{fig:exp_observable_nodes}, during the first 90~s, when both heater power and flow rates remained at zero, uncertainty was dominated by thermocouple noise in the inferred initial condition. As flow rates increased, uncertainty widened modestly due to advection effects; once heater power was applied, the uncertainty bands expanded substantially. 

As shown in Figure~\ref{fig:exp_observable_nodes}, the 95th-percentile ensemble bounds remain exceptionally well aligned with the measured trajectories at the observable nodes. This agreement provides strong empirical evidence that the discriminative fine-tuning methodology successfully adapted the temporal dynamics to the testbed while perfectly preserving the spatial, physics-consistent inductive biases acquired during SAM pretraining.

An additional, critical strength of the proposed model is the robust reconstruction of unmeasured states. Figure~\ref{fig:exp_missing_nodes} presents the inferred trajectories for the facility's permanently uninstrumented nodes, which remain smooth, physically plausible, and bounded by reasonable uncertainty margins. Because no direct measurements exist for these physical locations, ground-truth traces are unavailable for comparison; nevertheless, the stable latent trajectories confirm that the preserved energy-balance structure successfully regularizes the internal dynamics, preventing non-physical divergence even under noisy experimental boundary conditions. Confidence in these latent estimates is quantified using the same $M=64$ ensemble rollout procedure.

Additionally, to expand the range of operational data, including scenarios that cannot be safely reproduced experimentally, we require high-fidelity synthetic data generation. To this end, we will further improve the current SAM model, while Generative Artificial Intelligence (GenAI)-enabled modeling and simulation, combined with human expertise as demonstrated in prior studies \cite{NDUM2025100555, liu2025automatingdatadrivenmodelinganalysis, abulawi2026autosamagenticframeworkautomating}, will support this objective. 

\section*{Conclusions}
\label{sec:conclusions}

In this work, We presented a physics-informed Graph Neural ODE surrogate that couples message-passing spatial propagation with continuous-time latent dynamics for thermal-hydraulic forecasting under partial observability. On held-out SAM transients, the surrogate achieves an average MAE of 0.91~K at 60~s and 2.18~K at 300~s for uninstrumented nodes, with R$^2$ up to 0.995 for missing-node reconstruction. Inference runs at approximately 105$\times$ faster than simulated time on a single GPU, and 64-member ensemble rollouts remain practical for uncertainty quantification. To validate sim-to-real transfer, we adapted the SAM-pretrained model to experimental facility data using discriminative fine-tuning with only 30 training sequences. On a steep multi-step power transient ($0 \rightarrow 10 \rightarrow 5 \rightarrow 7$~kW), the surrogate tracked observable-node measurements within ensemble bounds and produced smooth, physically plausible trajectories at permanently uninstrumented locations.

These results demonstrate the GNN-ODE's potential as a forecasting engine for control-oriented digital twins in the reactor instrumentation and controls domain. The surrogate effectively functions as a virtual sensor array, reconstructing inaccessible thermal-hydraulic states from sparse measurements while maintaining the inference speed required for real-time supervisory loops. Its end-to-end differentiability is directly compatible with optimization-based frameworks such as Differentiable Predictive Control \cite{drgona2024}, and supports uncertainty-aware rollout analysis for safety-constrained decision making. The learned flow-dependent heat-transfer scaling recovered a Reynolds-number exponent consistent with the Dittus--Boelter correlation used in the reference SAM model, indicating that the surrogate learns interpretable constitutive structure beyond trajectory fitting. More broadly, when integrated with complementary supervisory frameworks such as Generative AI-assisted digital twin agents \cite{ndum2026llm}, the forecasting capability demonstrated here could support higher-level autonomous decision-making pipelines for advanced reactor operation.

Several limitations motivate future work, which we organize by priority. In the near term, we will close the multi-physics feedback loop by coupling the thermal-hydraulic GNN-ODE with a Point Kinetics Equation module, using the surrogate's hidden-state reconstruction to provide dynamic reactivity feedback for real-time reactor emulation. Additionally, deploying the surrogate for continual online adaptation will require robust noise-mitigation pipelines (e.g., recursive smoothing and signal filtering) to prevent sensor noise from destabilizing the continuous-time dynamics during retraining. In the medium term, we plan to evaluate topology transfer by instantiating the architecture on facilities with related but distinct thermal-hydraulic layouts, leveraging the general graph topology to test whether SAM-pretrained priors accelerate convergence on new configurations. Since the architecture natively supports dynamic node masking, we will also systematically study how selectively disabling sensor inputs for various fault combinations affects predictive accuracy across the graph. 

\section*{Acknowledgments}
This work is supported by the US Department of Energy Office of Nuclear Energy Distinguished Early Career Program under contract number DE-NE0009468.
The submitted manuscript has been co-created by UChicago Argonne, LLC, Operator of Argonne National Laboratory (“Argonne”). Argonne, a U.S. Department of Energy Office of Science laboratory, is operated under Contract No. DE-AC02-06CH11357.

\printbibliography

\end{document}